\documentclass{article} 
\usepackage{iclr2024_conference,times}
\usepackage{makecell}
\usepackage{tabularx}

\usepackage{amsmath,amsfonts,bm}

\def\eqref#1{equation~\ref{#1}}

\def\1{\bm{1}}

\DeclareMathAlphabet{\mathsfit}{\encodingdefault}{\sfdefault}{m}{sl}
\SetMathAlphabet{\mathsfit}{bold}{\encodingdefault}{\sfdefault}{bx}{n}

\usepackage{setspace}
\usepackage[hidelinks]{hyperref}
\usepackage{fontawesome}
\usepackage{xcolor}

\usepackage{graphicx}
\usepackage{eso-pic}

\hypersetup{
    pdftitle={From TeleChat to Xing: A Training Report on the Xing 4 Series},
    pdfsubject={Architecture and training systems for the Xing 4 series},
    colorlinks,
    linkcolor={magenta},
    citecolor={cyan},
    urlcolor={magenta}
}

\usepackage{url}
\usepackage{colortbl}

\usepackage{amsmath,bm}
\usepackage{graphicx}
\usepackage{microtype}
\usepackage{amsfonts}
\usepackage[noend]{algpseudocode}
\usepackage{algorithmicx,algorithm}
\usepackage{color}
\usepackage{etoolbox,lipsum}
\usepackage{float}
\usepackage{caption}
\usepackage{booktabs}
\usepackage{multirow}
\usepackage{array}
\usepackage{xspace}
\usepackage{enumerate}
\usepackage{nth}
\usepackage{cleveref}
\usepackage{bbm}
\usepackage{longtable}
\usepackage{CJKutf8}
\usepackage{tcolorbox}
\usepackage{xcolor}
\tcbuselibrary{breakable}

\title{
{
\centering
\fontsize{16}{19}\selectfont 
TeleChat to Xing4: Advancing Frontier \mbox{Agentic} Models on Domestic Chips and Frameworks
}
}

\author{
\parbox{\textwidth}    {\centering
    Xingchen AGI Lab, China Telecom
}%
}

\iclrfinalcopy 

\begin{document}

\AddToShipoutPicture*{%
  \AtPageUpperLeft{%
    \put(\LenToUnit{25mm},\LenToUnit{-20mm}){%
      \includegraphics[width=11mm]{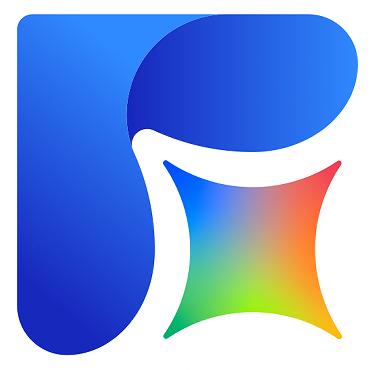}%
    }%
  }%
}

\maketitle

\begin{abstract}

We present Xing 4, a series of large language models that extends the TeleChat lineage through coordinated advances in model architecture and training infrastructure on Ascend. While most open-source models are trained on GPUs, scaling on domestic chips and frameworks poses significant systems challenges. This report describes the design principles and systems engineering behind the series, spanning architecture selection, large-scale pretraining, and reasoning and agentic post-training. We study how sparse activation, attention design, and model depth and width balance model capacity with computation, memory access, and execution overhead. The representative 285B configuration combines multi-head latent attention (MLA) with DeepSeek Sparse Attention (DSA) for long-context serving and adopts a wider, shallower mixture-of-experts (MoE) topology. We have open-sourced a 29B MoE member, a model trained entirely on Ascend NPUs using MindSpore, activating only 4B parameters per token while supporting native 256K-token contexts. Our pretraining infrastructure builds on MindSpore HyperParallel, combining topology-aware hybrid sharding, efficient distributed optimizer execution, and coordinated management of long-sequence workloads and activation memory. We combine fine-grained CPU affinity with core isolation to limit host scheduling interference. Together, these mechanisms support the pretraining of a 285B-parameter model on 8,192 Ascend 910C cards drawn from a large-scale production cluster, providing a concrete setting for the architectural and systems studies. In simulations at a context length of 256K, MLA with DSA and cross-layer index sharing (IndexShare) achieves 4.68 times the prefill throughput of MLA alone. At the same sequence length on 1,024 NPUs, measured step times at a fixed global batch size correspond to approximately 14.5\% higher training throughput for hybrid sharded data parallelism (HSDP) than for the 1F1B pipeline baseline. For post-training, we introduce TeleGym, a unified agent execution platform that connects execution, evaluation, and trajectory construction to policy optimization. Disaggregated asynchronous execution, bounded oversampling, and partial rollout accommodate heterogeneous task durations while preserving trajectory continuity across policy updates. Together, these advances establish an Ascend-oriented training foundation for the continued development of Xing 4 across model scales and increasingly demanding reasoning and agent workloads. The open-sourced 29B member and ongoing trillion-parameter training further extend the series within the Ascend and MindSpore ecosystem.

\end{abstract}

\newpage
\tableofcontents

\section{Introduction}
\label{sec:intro}

The TeleChat family has developed through successive advances in model capability and the infrastructure required to train at scale. TeleChat established an open foundation spanning pretraining, instruction tuning, and preference alignment~\citep{telechat,wang2024telechat}. TeleChat2, TeleChat2.5, and T1 expanded this foundation with stronger reasoning, coding, and long-context capabilities~\citep{wang2025telechat2}. MoE paradigm has rapidly evolved into the prevailing architectural choice for scaling large language models~\cite{hu2026awakening, hu2026mosaic}. TeleChat3-MoE extended the family to models with hundreds of billions to over a trillion parameters, accompanied by systematic work on numerical verification, distributed execution, and cluster performance~\citep{telechat3moe2025}. Throughout this progression, development of the models has been supported by sustained investment in the domestic Ascend and MindSpore ecosystem.

Building on this foundation, \textbf{Xing 4} continues the TeleChat lineage as a series spanning tens to hundreds of billions of parameters, with ongoing expansion toward trillion-parameter scale. The series targets long-context reasoning, code generation, and agentic applications while advancing efficient training on Ascend. Its members adopt architectural configurations suited to their parameter budgets and execution requirements.

We have open-sourced \textbf{Xing4.0-29B-A4B}, a MoE member with 29B total parameters, 4B activated parameters per token, and native support for 256K-token contexts~\citep{xing4029b}. Its architecture combines Manifold-Constrained Hyper-Connections (mHC)~\citep{xie2026mhc} with Multi-head Latent Attention (MLA) and multi-token prediction (MTP), bringing sparse computation and long-context processing to a model intended for reasoning and agentic use.

This focus on execution efficiency reflects a broader direction in foundation-model development. Qwen3 and Tencent's Hunyuan-A13B couple model capability with control over reasoning cost and deployment efficiency~\citep{qwen3,hunyuan2026a13b}. Kimi K3 and GLM-5 emphasize long-horizon reasoning and agentic execution~\citep{kimi2026k3,glm2026five}, while DeepSeek-V4 makes efficient million-token attention a central architectural objective~\citep{deepseek2026v4}. As documents, tool outputs, and intermediate reasoning accumulate, efficient execution becomes necessary both within model calls and across post-training trajectories.

Developing Xing 4 across parameter scales requires coordinated architectural and systems design. Increasing capacity must be balanced against per-token computation and the cost of storing, communicating, and updating the full model. Long-context attention adds indexing, KV-access, and activation-memory costs; packed documents can produce unequal computational loads even when token counts match. Host scheduling delays can stall device execution, while variable task durations introduce further imbalance during reinforcement learning. These constraints motivate joint control of model-state placement, computation ownership, and work scheduling.

At the architectural level, we study how model capacity and long-context requirements translate into inference costs. MLA compresses the KV cache~\citep{deepseek-v2}, DSA restricts each query to a sparse set of historical keys~\citep{deepseekai2025deepseekv32pushingfrontieropen}, and IndexShare reuses the selection across layers~\citep{bai2026indexcacheacceleratingsparseattention,zai2026glm52}. We also compare network depths and expert widths at similar parameter budgets to quantify the trade-off between matrix utilization and layer-wise routing and scheduling overhead. MTP provides an auxiliary objective during training~\citep{deepseek-v3}.

Our pretraining infrastructure is built on MindSpore HyperParallel~\citep{hyperparallel}. Topology-aware hybrid sharded data parallelism (HSDP), building on fully sharded data parallelism (FSDP)~\citep{zhao2023pytorch}, aligns model-state distribution with the cluster's bandwidth hierarchy and enables execution without pipeline stages. Owner-based Muon execution assigns full-matrix orthogonalization to designated processes within supported shard domains, avoiding redundant computation while preserving sharded state~\citep{liu2025muon}. For long sequences, cost-based workload assignment, reusable token layouts, and document-aware KV compaction address imbalance and memory access. Early indexer backward and selective recomputation control activation lifetimes. CPU affinity and core isolation reduce host contention that can delay task submission and collective participation.

For reasoning and agentic post-training, we introduce TeleGym, a unified execution platform for heterogeneous agent harnesses, environments, and persistent sessions. An intermediate server connects task interactions and evaluated outcomes to policy optimization through a shared trajectory representation. Task synthesis and validation reuse the same execution interfaces. Generation and optimization run asynchronously on separate device pools, with bounded oversampling accommodating variable task durations. Partial rollout allows unfinished trajectories to resume after policy updates. Sampling-time policy records and execution-level reward grouping preserve the information needed for learning as trajectories span multiple policy versions. Together, these mechanisms support mathematical reasoning, code generation, and multi-turn agent tasks within a common reinforcement-learning workflow.

The report makes three main contributions:

\begin{itemize}
    \item \textbf{Ascend-oriented architectural design for efficient training and inference.} We evaluate attention variants that reduce KV traffic and index computation at long context lengths. At comparable parameter budgets, we measure how a shallower network with wider expert layers affects training throughput and inference latency on Ascend.
    \item \textbf{Coordinated systems and cluster optimization for large-scale pretraining.} We develop a training infrastructure that combines topology-aware hybrid sharding and owner-based Muon execution with long-sequence workload balancing, KV compaction, and activation-memory management. CPU affinity and core isolation extend the optimization to host scheduling, coordinating device and host execution within MindSpore HyperParallel.
    \item \textbf{TeleGym and unified reinforcement learning for reasoning and agents.} We introduce TeleGym and connect its execution interfaces, task synthesis, and trajectory construction to a common policy-optimization pipeline. Disaggregated asynchronous execution, bounded oversampling, and partial rollout accommodate heterogeneous task durations while maintaining session continuity, sampling-time policy records, and execution-level reward grouping across policy updates.
\end{itemize}

The detailed engineering studies use a \textbf{285B-parameter MoE member} as the principal implementation example. It activates \textbf{17B parameters per token}, supports \textbf{1M-token contexts}, and is pretrained on \textbf{8,192 Ascend 910C cards}. The report combines architectural simulation, pretraining measurements, and operator-level analysis with a description of the post-training infrastructure; each evaluation specifies its model and hardware configuration.

Section~\ref{sec:model_architecture} presents the architectural design principles and a representative configuration, and Section~\ref{sec:training_optimization} details distributed pretraining. Section~\ref{sec:post_train_optimization} describes the reinforcement-learning framework for reasoning and agentic post-training. Section~\ref{sec:cluster_optimization} examines host scheduling and cluster throughput, and Section~\ref{sec:conclusion} concludes the report.

\section{Model Architecture and Design Principles}
\label{sec:model_architecture}

Architectural design for Xing 4 brings model capacity, long-context execution, and hardware efficiency into a common set of trade-offs. Agentic workloads such as long-context reasoning, multi-turn tool use, complex task planning, and high-concurrency inference~\citep{zhao2025enhancing,zhang2024lemur,zou2026group,bianagentvocab,li2026memory,xiong2025tablezoomer} make these requirements interdependent: longer histories increase attention cost, while repeated model calls expose decoding latency and sequential overhead.

Architectural choices vary with the parameter budget. The released 29B member combines mHC for residual-stream mixing~\citep{xie2026mhc} with MLA and MTP~\citep{xing4029b}. For the larger member, the studies below examine two decisions that affect the cost of serving long sequences on Ascend. One concerns the number of historical keys each query accesses. The other concerns the amount of computation in each layer.

\begin{figure}[htbp]
    \centering
    \includegraphics[width=0.9\linewidth]{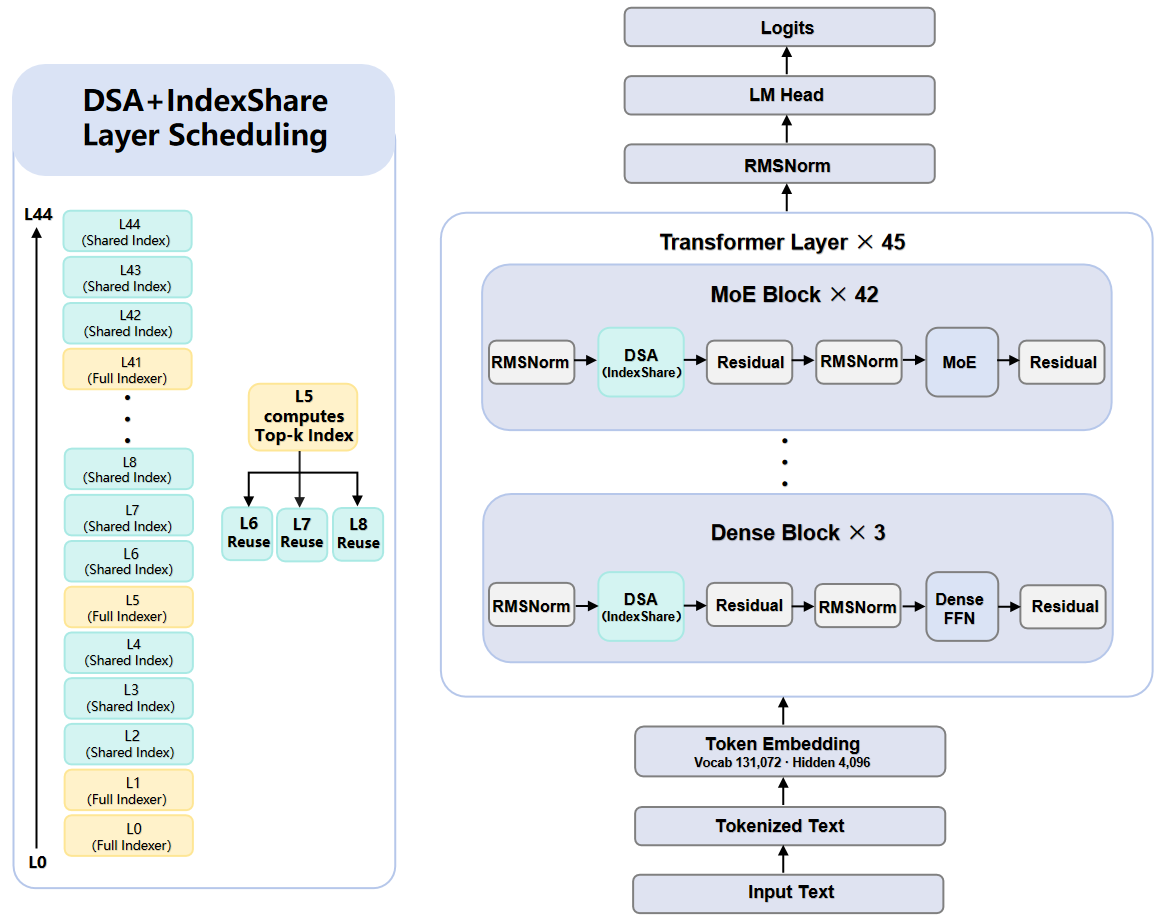}
    \caption{Architecture overview of the 285B representative model in the Xing 4 series.}
    \label{fig:xing4-285b-arch}
\end{figure}

Figure~\ref{fig:xing4-285b-arch} shows the final 285B architecture. The current preview uses MLA~\citep{deepseek-v2} with DSA~\citep{deepseekai2025deepseekv32pushingfrontieropen} and computes indices separately in each DSA layer. A subsequent 285B release will enable IndexShare~\citep{bai2026indexcacheacceleratingsparseattention,zai2026glm52}, reusing the selected indices across layers as shown in the figure. Of its 45 layers, the first three use dense feed-forward networks and the remaining 42 use sparsely activated MoE feed-forward networks.

From a hardware perspective, Ascend 910C provides strong Cube matrix-computation capability, while Vector compute and HBM bandwidth are relatively more constrained. These hardware characteristics guide our architectural choices:

\begin{quote}
\textbf{Leverage Cube with matrix-dominant computation, mitigate Vector bottlenecks through structural optimization, and reduce HBM pressure through sparse memory access.}
\end{quote}

\subsection{Design Objectives}

We select model configurations by balancing total capacity, per-token computation, and execution cost on Ascend 910C. Parameter budgets determine the available design space, while long-context requirements and hardware characteristics guide the choice of attention, expert configuration, and network depth. The aim is to preserve sufficient capacity while improving long-context deployability and end-to-end execution efficiency.

The architecture mainly targets two bottlenecks:

\begin{enumerate}
    \item \textbf{Long-context inference bottleneck}: MLA reduces KV Cache size, but in ultra-long-context scenarios, the main attention still needs to access a large number of historical KV states. As a result, decode latency increases with context length.
    \item \textbf{Ascend hardware efficiency bottleneck}: Overly deep networks amplify non-matrix-computation overheads, including normalization, routing, indexing, scheduling, and data movement, which reduces effective Cube utilization on Ascend 910C.
\end{enumerate}

These observations motivate two architectural studies.

\begin{itemize}
    \item \textbf{Attention design}: combining MLA with DSA and IndexShare reduces historical KV access and repeated index computation in long-context inference.
    \item \textbf{Depth-width trade-off}: balancing layer count and expert width changes matrix-computation density and sequential overhead; controlled comparisons identify configurations suited to Ascend 910C.
\end{itemize}

We retain MTP as an auxiliary training objective and use Muon to optimize matrix-valued parameters. Section~\ref{sec:training_optimization} describes how we distribute the Muon update across devices.

\subsection{Long-Context Attention}

Long-context attention requires control over both KV storage and the work performed for each query. We examine MLA as the baseline and add DSA and IndexShare to address historical KV access and repeated indexing, yielding a \textbf{MLA + DSA + IndexShare} design.

MLA compresses KV representations and reduces KV-cache memory storage requirement. DSA uses a lightning indexer to select Top-$K$ historical KV states for each query, so that the main attention only attends to a sparse subset of historical context. IndexShare further reuses sparse indices across layers, reducing repeated indexer computation in DSA and making sparse attention more suitable for long-context and high-concurrency decode workloads.

To evaluate these choices using attention variants of the 285B model on Ascend 910C, we primarily use simulation-based performance modeling based on Huawei's Zhanlu simulation framework. The simulation models the major computation, memory-access, and communication characteristics of the target Ascend 910C deployment, and is used to compare relative architectural trends under controlled conditions. Since simulation does not fully capture all production-serving overheads, the reported results may be more optimistic than full end-to-end deployment measurements. The comparison covers three variants: MLA, MLA+DSA, and MLA+DSA+IndexShare, under a prefill/decode(P/D) disaggregation deployment. On the prefill side, we focus on long-prompt throughput and time to first token(TTFT); on the decode side, we focus on time per output token(TPOT) and concurrent decode throughput.


\begin{table}[htbp]
    \centering
    \caption{Inference evaluation setup for attention variants of the 285B model.}
    \label{tab:attention-eval-config}
    \renewcommand{\arraystretch}{1.3}
    \begin{tabular}{lc}
        \hline
        Item & Configuration \\
        \hline
        Inference environment & 32 Ascend 910C cards \\
        Deployment mode & P/D disaggregation \\
        Prefill-side parallelism & TP4DP4 \\
        Decode-side parallelism & TP2DP8 \\
        Evaluation context length & 256K \\
        \hline
    \end{tabular}
\end{table}

Table~\ref{tab:attention-arch-comparison} compares the three variants. All results are normalized against the MLA baseline.

\begin{table}[htbp]
    \centering
    \caption{Simulation-based inference efficiency of attention variants in the 285B study. All metrics are normalized against the MLA baseline. For access and latency metrics, lower is better; for throughput metrics, higher is better.}
    \label{tab:attention-arch-comparison}
    \footnotesize
    \renewcommand{\arraystretch}{1.3}
    \setlength{\tabcolsep}{4pt}
    \begin{tabular}{@{} l c c c c c c @{}}
        \toprule
        Architecture 
        & \makecell{KV HBM\\Access} 
        & \makecell{Index\\Computation} 
        & \makecell{Relative\\TTFT} 
        & \makecell{Relative\\TPOT} 
        & \makecell{Prefill\\Throughput} 
        & \makecell{Decode\\Throughput} \\
        \midrule
        MLA 
        & 1.00$\times$ 
        & None 
        & 1.00$\times$ 
        & 1.00$\times$ 
        & 1.00$\times$ 
        & 1.00$\times$ \\
        MLA + DSA 
        & 0.85$\times$ 
        & Per-DSA-layer 
        & 0.37$\times$ 
        & 0.84$\times$ 
        & 2.67$\times$ 
        & 1.18$\times$ \\
        MLA + DSA + IndexShare 
        & 0.80$\times$ 
        & Cross-layer 
        & 0.21$\times$ 
        & 0.79$\times$ 
        & 4.68$\times$ 
        & 1.27$\times$ \\
        \bottomrule
    \end{tabular}
\end{table}

In the 256K-context simulation, the MLA + DSA variant delivers 2.67 times the prefill throughput of MLA alone. With IndexShare, the ratio rises to 4.68. The corresponding decode throughput ratios are 1.18 and 1.27. Under this workload, the larger benefit of cross-layer index reuse is in prefill.

\subsection{Depth-Width Trade-offs}

Model depth and expert width jointly determine the balance between matrix computation and layer-wise sequential work. In Xing 4 architecture selection, we consider the number of layers, hidden size, expert intermediate size, and the number of activated experts per token together, seeking configurations that fit both the parameter budget and Ascend 910C execution characteristics.

Deeper networks increase layer-wise sequential overhead, especially in non-matrix-computation components such as normalization, routing, and indexing. In contrast, a wider-and-shallower topology increases the scale of per-layer matrix multiplications, improves Cube utilization, and reduces cross-layer scheduling and communication overhead.

The released 29B member provides one concrete configuration: 40 layers, 64 routed experts with four selected per token, and one shared expert~\citep{xing4029b}. At larger parameter budgets, depth and expert width must be reconsidered together. The following controlled study compares candidate topologies for the 285B model while holding the main architectural components fixed.

Our depth-width study compares a baseline topology with a wider, shallower alternative. Both configurations share the same base architecture, consisting of MLA, DSA, IndexShare, a hybrid Dense/MoE backbone, and an MTP objective. The main differences are the number of layers, expert intermediate size, and the number of activated experts per token.

\begin{table}[htbp]
    \centering
    \caption{Candidate configurations for the representative model's depth-width study.}
    \label{tab:depth-width-candidates}
    \small
    \renewcommand{\arraystretch}{1.3}
    \setlength{\tabcolsep}{3.5pt} 
    \begin{tabular}{lccccccc}
        \toprule 
        Configuration &
        \makecell{Layers} &
        \makecell{Hidden\\Size} &
        \makecell{Heads} &
        \makecell{Routed\\Experts} &
        \makecell{Expert\\Intermediate\\Size} &
        \makecell{Experts\\per Token} &
        \makecell{Shared\\Experts} \\
        \midrule 
        Baseline        & 60 & 4096 & 128 & 256 & 1536 & 6 & 1 \\
        Wider-shallower & 45 & 4096 & 128 & 256 & 2048 & 8 & 1 \\
        \bottomrule 
    \end{tabular}
\end{table}

The inference-side comparison focuses on decode latency and throughput under long-context settings. Results are normalized against the baseline configuration.

\begin{table}[htbp]
    \centering
    \caption{Normalized inference efficiency in the 285B depth-width study.}
    \label{tab:depth-width-inference}
    \small
    \renewcommand{\arraystretch}{1.3}
    \setlength{\tabcolsep}{4.5pt}
    \begin{tabular}{@{} l c c c c c @{}}
        \toprule
        Configuration 
        & \makecell{Relative\\TTFT} 
        & \makecell{Relative\\TPOT} 
        & \makecell{Prefill\\Throughput} 
        & \makecell{Decode\\Throughput} 
        & Decision \\
        \midrule
        Baseline 
        & 1.00$\times$ 
        & 1.00$\times$ 
        & 1.00$\times$ 
        & 1.00$\times$ 
        & Reference \\
        Wider-shallower 
        & 0.83$\times$ 
        & 0.90$\times$ 
        & 1.20$\times$ 
        & 1.11$\times$ 
        & Selected \\
        \bottomrule
    \end{tabular}
\end{table}

Simulation and real-system measurements show that, under comparable total and activated parameter budgets, the 45-layer wider-FFN topology provides better inference latency and throughput than the 60-layer narrower-FFN topology. Large-scale training measurements also show that the wider-and-shallower topology improves overall training throughput by approximately 10\% compared with the deeper baseline.

We therefore select the 45-layer wider-FFN topology. This design preserves model scale and quality while reducing cross-layer sequential overhead and increasing the proportion of matrix-dominant computation on Ascend 910C, making it more suitable for long-context, high-concurrency inference and large-scale training.

\subsection{Representative Model Configurations}

Table~\ref{tab:xing4-model-configs} summarizes two members of the Xing 4 series with different parameter budgets and architectural configurations. The 285B member serves as the principal implementation for the architectural and training-system studies in this report.

\begin{table}[htbp]
    \centering
    \caption{Representative configurations of the Xing 4 series. The per-token expert count includes selected routed experts only; shared experts are listed separately.}
    \label{tab:xing4-model-configs}
    \small
    \renewcommand{\arraystretch}{1.3}
    \setlength{\tabcolsep}{4pt}
    \begin{tabular}{lcccccc}
        \toprule
        Model & Layers & \makecell{Hidden\\Size} & \makecell{Routed\\Experts} & \makecell{Expert\\Intermediate\\Size} & \makecell{Experts\\per Token} & \makecell{Shared\\Experts} \\
        \midrule
        \textbf{Xing 4 (29B)} & 40 & 3584 & 64 & 1024 & 4 & 1 \\
        \textbf{Xing 4 (285B)} & 45 & 4096 & 256 & 2048 & 8 & 1 \\
        \bottomrule
    \end{tabular}
\end{table}

Xing 4 (29B) corresponds to the open-sourced Xing4.0-29B-A4B. It activates 4B parameters per token, natively supports 256K-token contexts, and combines mHC with MLA and MTP~\citep{xing4029b}.

The 285B member activates approximately 17B parameters per token and supports contexts of up to one million tokens. Its 45-layer backbone begins with three dense FFN layers followed by 42 MoE layers. The final attention design combines MLA, DSA, and IndexShare. We train with an auxiliary MTP objective and optimize matrix-valued parameters with Muon. This configuration meets the design budgets of fewer than 300B total parameters and fewer than 20B activated parameters per token.

\section{Training Infrastructure and Optimizations in HyperParallel}
\label{sec:training_optimization}

The Xing 4 pretraining system uses MindSpore HyperParallel to shard model state across a two-tier Ascend network. High-bandwidth links within each SuperPoD carry weight gathers and sharded-gradient collectives. Gradient reductions across replicas use the links between SuperPoDs. This layout supports training without pipeline stages. The following sections examine sharded communication, distributed Muon updates, selective recomputation, and sparse attention for long sequences.

We profiled the 285B configuration on 1,024 NPUs. Production pretraining used 8,192 Ascend 910C cards. In the 256K-token step-time comparisons, HSDP reduced the PP-1F1B baseline from 36.4 to 31.8 s. With HSDP fixed, assigning each Muon matrix to one owner reduced step time to 28.9 s (9.1\%). Two independent experiments then used 28.9 s as their baseline. Selective recomputation reduced step time to 26.9 s (6.9\%), and layer-wise indexer backward reduced it to 24.5 s (15.2\%).


\subsection{Hybrid Sharded Data Parallelism (HSDP)}
\label{sec:hyperparallel-fsdp}

Training mixture-of-experts (MoE) models with hundreds of billions of parameters typically combines data, tensor, expert, and pipeline parallelism (DP/TP/EP/PP). While pipeline parallelism (PP) relieves memory pressure by partitioning model layers across stages, it inevitably introduces a structural pipeline bubble. For instance, with \(p=4\) pipeline stages and \(m=8\) micro-batches, the theoretical bubble fraction of the 1F1B schedule is \(\frac{p-1}{m+p-1}=\frac{3}{11}\approx 27\%\). Although interleaved schedules, such as virtual pipeline parallelism (VPP), can mitigate this issue, they cannot eliminate the pipeline bubble entirely.

The primary benefit of PP lies in its layer-wise memory partitioning, which circumvents per-device memory bottlenecks. If the data-parallel dimension could instead provide this memory relief---ensuring that no single accelerator needs to keep the full parameter tensors of all layers resident simultaneously---PP would become unnecessary, thereby eliminating the pipeline bubble altogether.

Fully Sharded Data Parallel (FSDP) \cite{zhao2023pytorch} adopts this philosophy. Model parameters are stored as distributed shards. Before each layer is executed, an AllGather operation reconstructs its full parameters; after the computation, the parameters are resharded and the temporary full tensors are released. This approach enables large-scale model training with minimal parameter redundancy.

However, when FSDP is scaled to clusters containing thousands or tens of thousands of NPUs, the sharding group may grow proportionally with the cluster size. Consequently, frequent AllGather and ReduceScatter operations must cross SuperPoD boundaries. These fine-grained collectives incur high communication latency, underutilize the relatively limited inter-SuperPoD bandwidth, and are difficult to overlap completely with computation. Moreover, the additional memory savings obtained from an excessively large sharding degree exhibit diminishing returns.

\begin{figure}[t]
  \centering
  \includegraphics[width=0.95\textwidth] {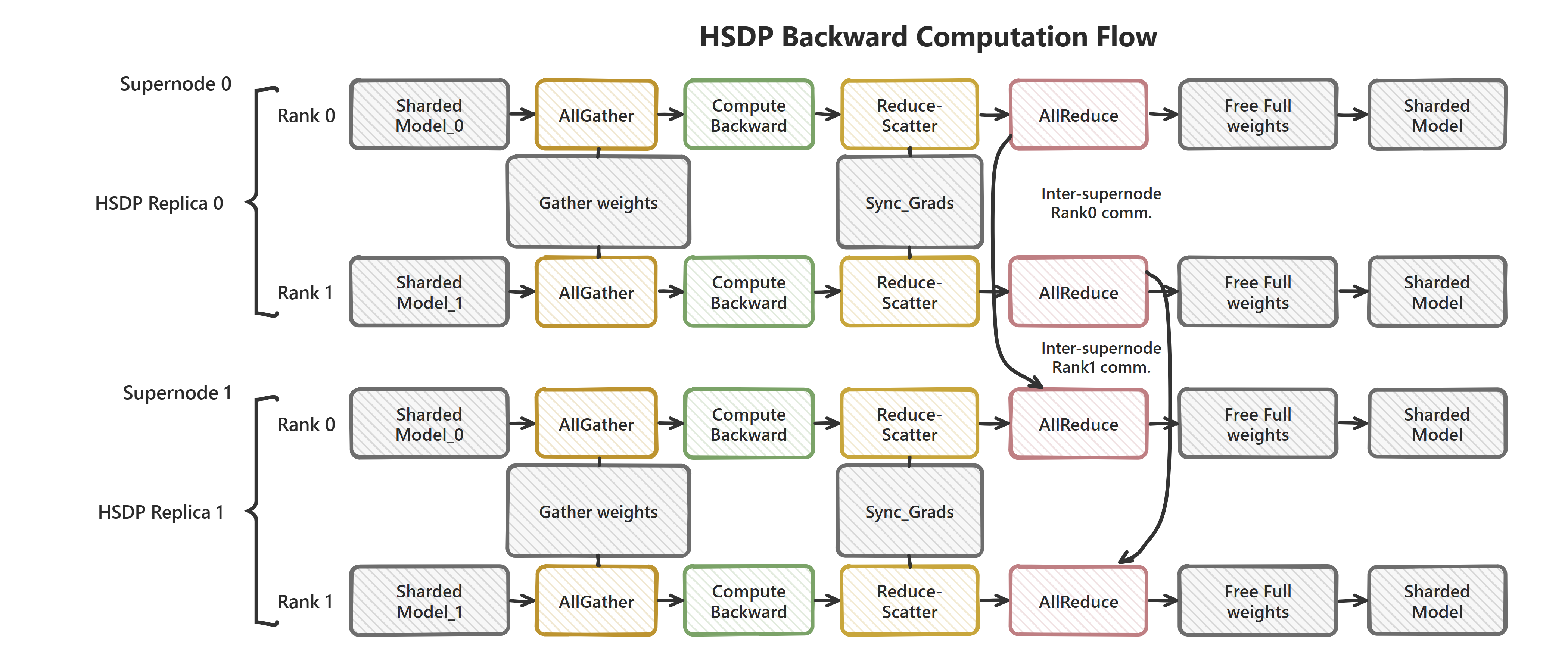} \caption{Backward-pass workflow of HyperParallel-FSDP.} \label{fig:backward-workflow}
\end{figure}

To address these challenges, we design and implement \textbf{HyperParallel-FSDP} \cite{hyperparallelfsdp2026,mindspore-framework}, a hybrid sharded data-parallel (HSDP) scheme tailored to the Ascend SuperPoD architecture. Model parameters are FSDP-sharded within each SuperPoD and replicated across SuperPoDs. Frequent weight-gathering and sharded-gradient synchronization operations are therefore confined to the high-bandwidth intra-SuperPoD communication domain, while only coarse-grained gradient reduction traffic crosses SuperPoD boundaries. This design balances memory efficiency and communication overhead while enabling the removal of pipeline parallelism. The backward-pass workflow of HyperParallel-FSDP is illustrated in Figure~\ref{fig:backward-workflow}.

The Ascend SuperPoD provides ultra-high intra-SuperPoD bandwidth and ultra-low intra-SuperPoD communication latency. HyperParallel-FSDP introduces two key designs that exploit these architectural properties.

\begin{figure}[t]
  \centering
  \includegraphics[width=0.95\textwidth] {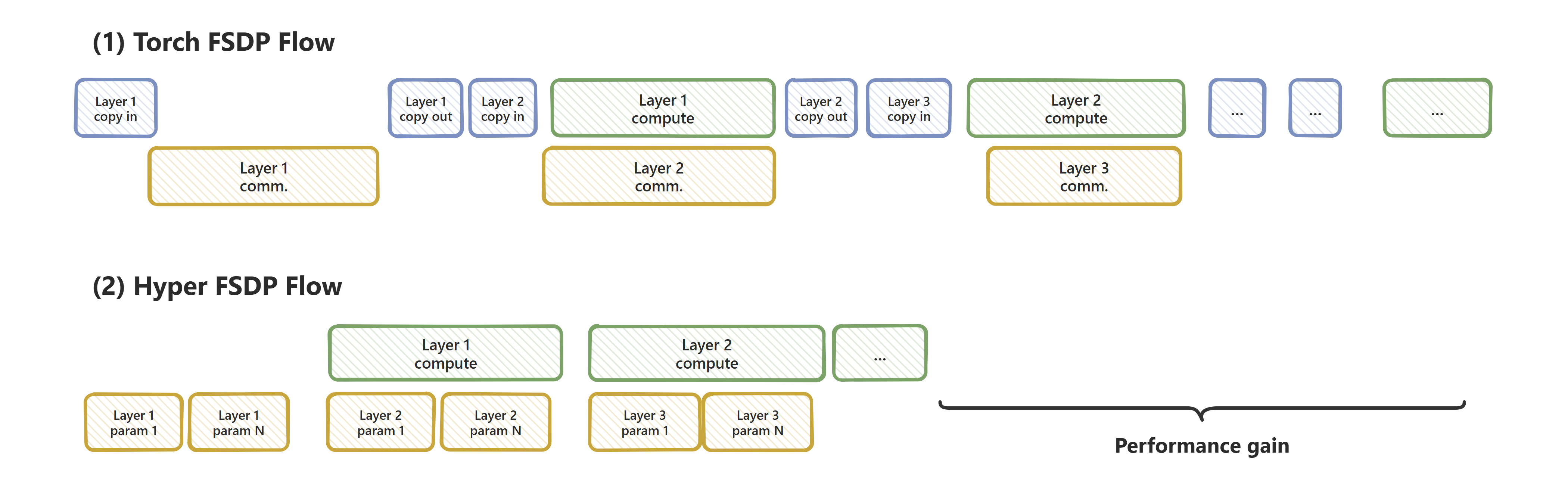} \caption{(1) Fused communication in PyTorch FSDP and (2) zero-copy communication in HyperParallel-FSDP.} \label{fig:zero-copy}
\end{figure}

The first design targets the additional data movement introduced by parameter flattening. To improve collective communication efficiency, PyTorch FSDP \cite{zhao2023pytorch} flattens multiple parameters into a single contiguous buffer and then performs a fused collective operation, as shown in Figure~\ref{fig:zero-copy}(1). Although this design reduces the number of collective operations, it introduces additional data movement. Before communication, individual parameter tensors must be copied into the flattened communication buffer (\emph{copy-in}). After AllGather, the reconstructed parameters must be copied from the communication buffer back to their corresponding parameter tensors (\emph{copy-out}). Copy-out lies on the critical execution path and cannot be overlapped with subsequent computation because the computation directly depends on the gathered parameters.

The Ascend SuperPoD provides sufficiently high intra-SuperPoD bandwidth for fine-grained collective operations to retain high communication efficiency. HyperParallel-FSDP therefore performs AllGather and ReduceScatter directly on individual parameter tensors, as illustrated in Figure~\ref{fig:zero-copy}(2). This per-parameter communication design removes the flattened intermediate buffer and eliminates the associated parameter packing, copy-in, and copy-out operations. Consequently, HyperParallel-FSDP achieves \textbf{zero-copy communication} while preserving efficient intra-SuperPoD collective communication.

The second design fuses inter-SuperPoD gradient reduction and overlaps multiple communication operations with backward computation. During backpropagation, the execution of each layer involves four major operations: intra-SuperPoD AllGather for parameter reconstruction, backward computation, intra-SuperPoD ReduceScatter for sharded-gradient synchronization, and inter-SuperPoD AllReduce for gradient reduction across replicas.

\begin{figure}[t]
  \centering
  \includegraphics[width=0.98\textwidth] {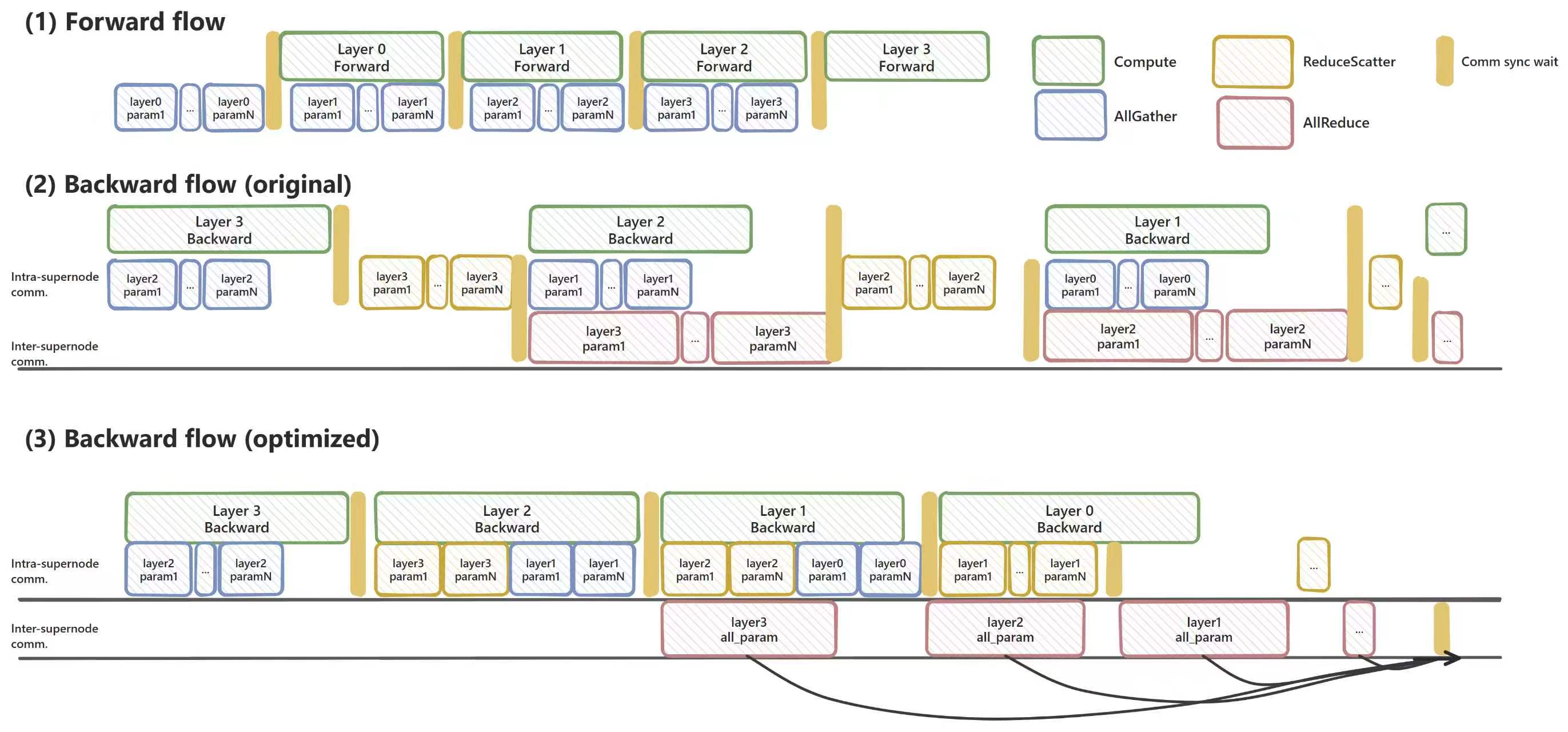} \caption{Execution timelines of HyperParallel-FSDP: (1) forward pass, (2) baseline backward pass, and (3) optimized backward pass with three-way communication overlap.} \label{fig:communication-overlap}
\end{figure}

Figure~\ref{fig:communication-overlap}(1) shows the forward-pass execution timeline. Before the forward computation of each layer, the required parameter shards are reconstructed through an intra-SuperPoD AllGather. Once the full parameters become available, the corresponding layer computation can proceed.

A straightforward backward implementation is shown in Figure~\ref{fig:communication-overlap}(2). This baseline schedule overlaps part of the inter-SuperPoD AllReduce communication with backward computation, but it has two limitations. First, frequent fine-grained AllReduce operations are inefficient because inter-SuperPoD bandwidth is more limited than intra-SuperPoD bandwidth. Second, intra-SuperPoD ReduceScatter and communication synchronization remain partially exposed on the critical execution path, resulting in visible communication gaps between consecutive layer computations.

HyperParallel-FSDP addresses these limitations through three complementary techniques:

\begin{enumerate}
 
  \item Intra-SuperPoD AllGather and ReduceScatter operations are asynchronously scheduled and overlapped with backward computation.
  \item Inter-SuperPoD gradient reduction proceeds asynchronously once the corresponding gradient data become available. Synchronization is deferred until the end of the backward pass, allowing its latency to be hidden behind the backward computation of multiple layers.
  \item Gradient shards participating in inter-SuperPoD reduction are aggregated and communicated through fused AllReduce operations, reducing communication frequency and collective-launch overhead. \end{enumerate}

To satisfy the contiguous-memory requirement of fused AllReduce, HyperParallel-FSDP pre-allocates a contiguous gradient buffer and configures each ReduceScatter operation to write its output directly into the corresponding region of this buffer. The gradient shards are therefore contiguous by construction, so the subsequent fused AllReduce requires no additional packing or memory-copy operations.

The optimized schedule is shown in Figure~\ref{fig:communication-overlap}(3). During the backward computation of layer \(i\), the system can simultaneously overlap the ReduceScatter of layer \(i+1\), the AllGather of layer \(i-1\), and inter-SuperPoD AllReduce communication. Communication synchronization is postponed until the end of the backward pass. This arrangement achieves \textbf{three-way communication overlap}, substantially reducing exposed communication latency.

To evaluate the effectiveness of the proposed design, we compare conventional pipeline-parallel configurations with HyperParallel-FSDP on the 1,024-NPU profiling cluster, using a 285-billion-parameter MoE model. The sequence length is set to \(262{,}144\), and the global batch size is \(16\). Tensor parallelism (TP), expert parallelism (EP), and context parallelism (CP) are fixed at \(4\), \(64\), and \(32\), respectively. Table~\ref{tab:performance} summarizes the experimental results.

\begin{table}[htbp]
  \centering
  \caption{Training performance of the 285B model on the 1,024-NPU profiling cluster.} \label{tab:performance}
  \small

\begin{tabular}{lccccc}
  \toprule \textbf{Scheme} & \textbf{Pipeline} & \textbf{DP (Replica \(\times\) Shard)} & \textbf{mbs} & \textbf{Step Time} & \textbf{Speedup} \\
  \midrule PP-1F1B (Baseline) & \(pp=4,\ vpp=1\) & \(2 \times 1\) & \(8\) & \(36.4\,\mathrm{s}\) & \(1.00\times\) \\
  PP-Interleaved & \(pp=4,\ vpp=2\) & \(2 \times 1\) & \(4\) & \(34.2\,\mathrm{s}\) & \(1.06\times\) \\
  HSDP (\textbf{Ours}) & \(pp=1\) & \(4 \times 2\) & \(2\) & \(31.8\,\mathrm{s}\) & \(1.14\times\) \\
  \bottomrule \end{tabular}

 \end{table}

\noindent {\small\emph{Note:} \textsf{mbs} denotes the micro-batch size.}

At $PP=4$, interleaving the pipeline schedule reduced step time from 36.4 to 34.2 s (6.0\%) relative to 1F1B. The HSDP configuration used a single pipeline stage ($PP=1$) and distributed model state across four replicas with two shards per replica. With the same global batch size and sequence length, its step time was 31.8 s, a 12.6\% reduction relative to PP-1F1B. At this fixed workload, the ratio of step times corresponds to approximately 14.5\% higher training throughput.

The HSDP result in Table~\ref{tab:performance} reflects the full parallel configuration, including the data-parallel sharding layout, micro-batch size, and communication schedule. In this configuration, hybrid sharding fits the model within device memory and removes the pipeline bubble, while direct writes into the gradient buffer and three-way overlap limit exposed communication time.

\subsection{Owner-Based Execution for the Distributed Muon Optimizer}
\label{sec:muon-infra}

Muon orthogonalizes matrix-valued parameters with a Newton--Schulz (NS) iteration, which requires the full weight matrix. This conflicts with sharded training, where each rank holds only a slice of the parameters and optimizer state. In a straightforward distributed implementation, every rank that owns a slice of a two-dimensional weight updates its local momentum, reconstructs the full NS input, runs the same NS iteration, and keeps only the part of the resulting update that belongs to its local shard. The NS computation is therefore duplicated on every participating rank even though each rank uses only a fraction of the result. The redundancy lies in who performs the full-matrix step, not in how the state is sharded.

\begin{figure}[t]
  \centering
  \includegraphics[width=\linewidth]{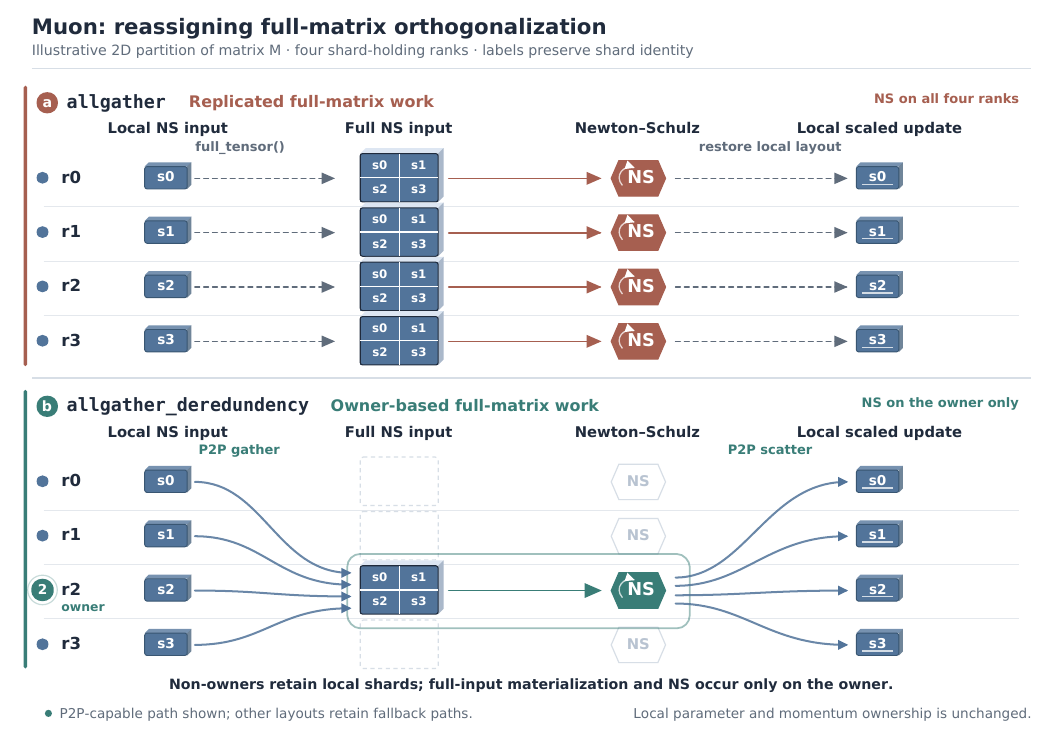}
  \caption{Reassignment of full-matrix Muon work. Four ranks hold illustrative shards of one matrix, with rank r2 selected as its owner. The baseline reconstructs and orthogonalizes the input on each participating rank. The supported P2P path reconstructs it at the owner and returns scaled-update shards to their original destinations. The owner's own input and output slices remain local. Faint outlines indicate omitted full-input materialization and NS, not missing local state. The shard arrangement is illustrative; arrows do not encode communication volume.}
  \label{fig:muon-owner}
\end{figure}

We remove this duplication by assigning each matrix a single owner rank within its shard group. Non-owner ranks send their input shards to the owner over point-to-point (P2P) transfers. The owner assembles the full input, runs NS, scales the update, keeps its own slice, and scatters the remaining slices back. Parameters and momentum stay in their original sharded layout; only the full-matrix computation moves to the owner (Figure~\ref{fig:muon-owner}). Ownership is per matrix rather than global, so replica-local shard groups can use different owners. Layouts that cannot use this P2P path fall back to the original execution path.

The scheduler groups matrices by shape, shard group, and estimated cost, then distributes the groups across eligible owners. It stacks compatible inputs for batched matrix multiplication to amortize kernel launches. Each matrix retains its own normalization and Newton--Schulz iteration. The distribution of matrix shapes and costs determines how evenly the owners share work and temporary storage.

\begin{figure}[t]
  \centering
  \includegraphics[width=\linewidth]{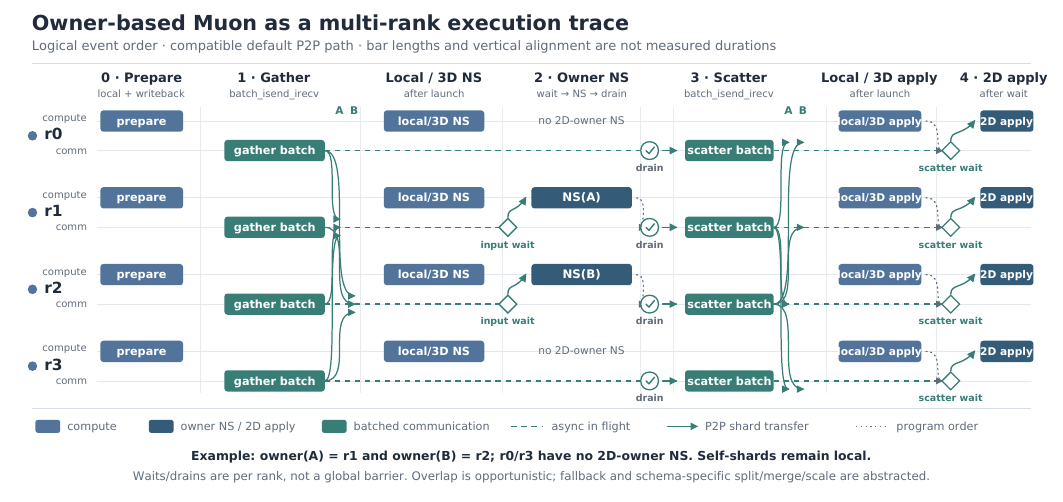}
  \caption{Logical execution order for a compatible owner-based P2P path. Matrix A is assigned to r1 and B to r2. Compute and communication lanes expose batched submission, input waits, gather draining, and scatter completion dependencies. Self-shards remain local. Local/3D work is shown where available; fallback paths and model-specific transformations are omitted. Bar lengths, dashed spans, and vertical alignment are schematic and do not represent measured durations or guaranteed overlap.}
  \label{fig:muon-execution}
\end{figure}

Figure~\ref{fig:muon-execution} shows where each rank waits for its data. After updating its local momentum, a rank submits batched gathers. Independent local or three-dimensional NS updates can run while an owner collects the shards it needs. Once the owner has assembled and orthogonalized the input, it scatters the result. A two-dimensional parameter update waits for its return shard. Communication handles track the batched transfers, allowing independent work to overlap with communication within a rank.

The implementation uses separate temporary-memory estimates to group preparation tasks, parameter updates, and NS work on local or expert matrices. Owners also allocate space for gathered two-dimensional inputs and their orthogonalization. Momentum is written back after local preparation, and full inputs can be released after NS. Peak memory depends on how long these buffers and any outstanding transfers remain live.

The owner-based P2P path applies to compatible shard layouts. Other layouts retain the original update path, and compatibility mode runs NS separately for each matrix. Per-matrix normalization and orthogonalization use the same matrix boundaries in both paths. Batched execution can change the order of floating-point operations.

We measured the effect of owner assignment with the HSDP layout in Table~\ref{tab:performance} held fixed. Changing the optimizer execution path reduced step time from 31.8 to 28.9 s (9.1\%, Table~\ref{tab:muon-perf}). Parameter and momentum shards kept their original placement. The owner performed the full-matrix NS computation once for each matrix in its shard group.

\begin{table}[htbp]
  \centering
  \caption{Effect of owner-based Muon execution under a fixed HSDP layout (\(1{,}024\) NPUs, sequence length \(262{,}144\)).}
  \label{tab:muon-perf}
  \small
  \begin{tabular}{lccc}
    \toprule
    \textbf{Configuration} & \textbf{Step Time} & \textbf{Speedup} & \textbf{Step-time reduction} \\
    \midrule
    HSDP & \(31.8\,\mathrm{s}\) & \(1.00\times\) & --- \\
    HSDP + owner-based Muon & \(28.9\,\mathrm{s}\) & \(1.10\times\) & \(9.1\%\) \\
    \bottomrule
  \end{tabular}
\end{table}

\subsection{Fine-Grained Selective Recomputation}
\label{sec:recompute}

Layer-wise recomputation offers only a coarse memory--compute trade-off: saving or replaying an entire layer treats all of its operators alike. This is particularly inefficient for long-sequence DSA, where some operators are expensive to execute but produce relatively small outputs. Replaying these operators incurs substantial computation for limited activation-memory savings. Native operator-level selective checkpointing provides finer control through a dispatch mode, but every operator dispatch invokes a Python policy callback. This adds host work throughout the forward pass and can make execution host-bound when device kernels are short. Policies based on operator type also make it cumbersome to mark repeated instances of the same operator differently within a layer.

\subsubsection{Object-Based Exclusion and Memory-Budgeted Selection}

Our \emph{exclude} scheme provides operator-level recomputation control through named model objects. An operator is assigned as an object member during initialization, and its member path can then be configured to select whether it is recomputed. Separate members distinguish different uses of the same operator type. Expensive operators are placed in exclude regions, while the surrounding computation remains eligible for replay. This does not require a framework-wide dispatch policy or a Python policy callback at every operator dispatch, preserving fine control while reducing host-side callback overhead.

\begin{figure}[htbp]
  \centering
  \includegraphics[width=\linewidth]{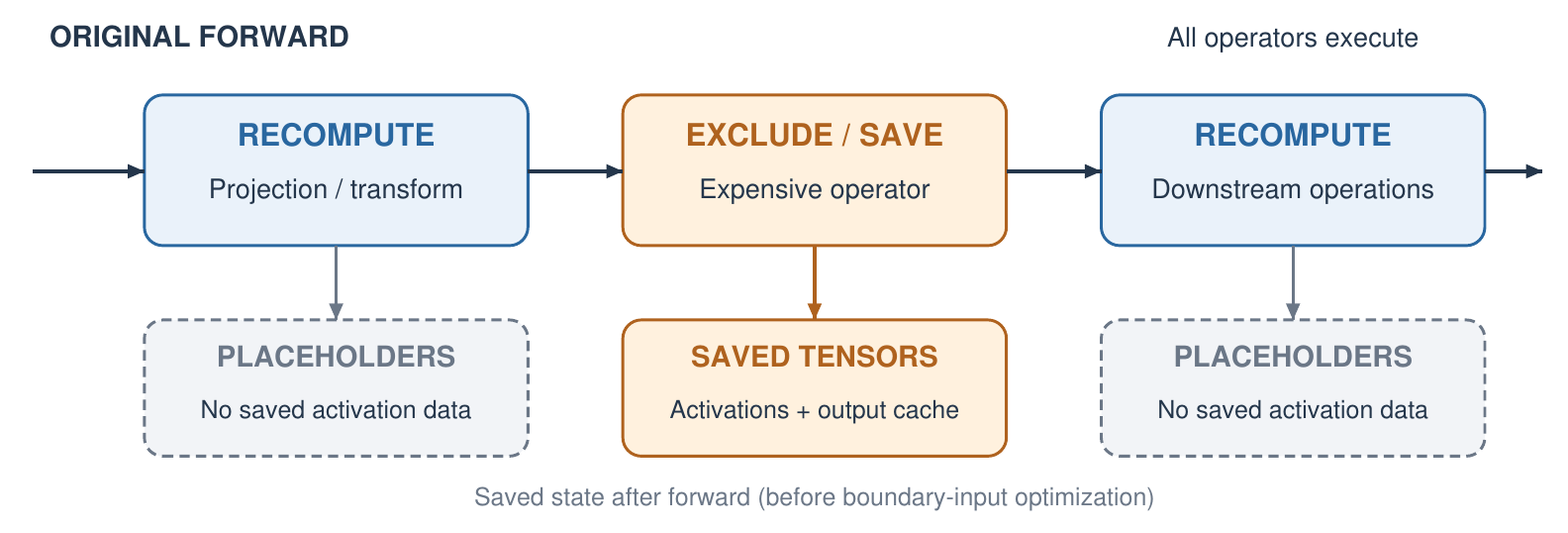}
  \caption{Activation handling in the original forward pass. Blue regions replace saved activations with placeholders; the orange exclude region retains activations and its output. Both regions execute normally in the forward pass.}
  \label{fig:recompute-flow}
\end{figure}

The forward pass has two kinds of activation regions, illustrated in Figure~\ref{fig:recompute-flow}. In a recompute region, activations that would otherwise be saved for backward are replaced by placeholders. In an exclude (save) region, the corresponding activations and operator outputs are retained. During backward replay, the former activations are regenerated and the excluded operators reuse their saved outputs. The choice therefore avoids replaying selected expensive operators without retaining every intermediate activation in their surrounding layer.

We choose excluded operators from profiling data. For each candidate operator $i$ we record its forward execution time $t_i$ together with its output shapes and data types, from which we estimate the output-memory cost $m_i$. Let $M$ be the activation-memory budget remaining after the other training allocations, and let $x_i=1$ mean that the operator output is retained and the operator is excluded from replay. Selection is then a 0--1 knapsack problem:
\begin{equation}
  \max_{x_i\in\{0,1\}}\;\sum_i x_i t_i
  \qquad\text{subject to}\qquad
  \sum_i x_i m_i\leq M.
  \label{eq:recompute-knapsack}
\end{equation}
Retaining an output costs $m_i$ bytes and removes $t_i$ of replay computation; an operator that is not selected stays eligible for replay. The objective therefore spends a bounded amount of activation memory on the operators that are most expensive to execute, instead of applying one policy to an entire layer.

We profiled 150 compute kernels in the forward pass of one Transformer layer on rank 0 at a context length of 256K. Their durations summed to 115.1 ms. Table~\ref{tab:recompute-kernels} lists selected kernel types and their logical output sizes. SparseLightningIndexerGradKLLoss accounted for 38.32 ms and 193 MiB, AllGatherMatmul for 15.90 ms and 656 MiB, and SparseFlashAttention for 13.47 ms and 516 MiB. The different orderings by time and output size illustrate why the selection rule accounts for both replay cost and the memory needed to retain an output.

\begin{table}[htbp]
  \centering
  \caption{Compute kernels of one forward Transformer layer on rank~0 (256K-token run), aggregated by kernel name. Output size is the logical total derived from the output shapes and data types, not peak device memory.}
  \label{tab:recompute-kernels}
  \small
  \begin{tabular}{lrr}
    \toprule
    \textbf{Kernel} & \textbf{Duration (ms)} & \textbf{Output size (MiB)} \\
    \midrule
    SparseLightningIndexerGradKLLoss & 38.32 & 193.0 \\
    LightningIndexer & 21.51 & 48.0 \\
    AllGatherMatmul & 15.90 & 656.0 \\
    SparseFlashAttention & 13.47 & 516.0 \\
    MatmulReduceScatter & 6.00 & 64.0 \\
    \bottomrule
  \end{tabular}
\end{table}

Under the profiled memory budget, Equation~\ref{eq:recompute-knapsack} selects the set of outputs that saves the most replay time in aggregate. Other operators remain eligible for replay. At the boundary of each save region, we also release inputs that backward can recover from the adjoining recompute region (Section~\ref{sec:recompute-boundary}).

With HSDP and owner-based Muon enabled, selective recomputation reduced step time from 28.9 to 26.9 s (6.9\%, Table~\ref{tab:recompute-perf}). This comparison measures the end-to-end effect of the recomputation policy at the same sequence length and on the same devices.

\begin{table}[htbp]
  \centering
  \caption{Effect of fine-grained selective recomputation, evaluated against the HSDP\,+\,Muon configuration (\(1{,}024\) NPUs, sequence length \(262{,}144\)).}
  \label{tab:recompute-perf}
  \small
  \begin{tabular}{lccc}
    \toprule
    \textbf{Configuration} & \textbf{Step Time} & \textbf{Speedup} & \textbf{Step-time reduction} \\
    \midrule
    HSDP + Muon & \(28.9\,\mathrm{s}\) & \(1.00\times\) & --- \\
    HSDP + Muon + selective recomputation & \(26.9\,\mathrm{s}\) & \(1.07\times\) & \(6.9\%\) \\
    \bottomrule
  \end{tabular}
\end{table}

\subsubsection{Reducing Activation Memory Inside Exclude Regions}
\label{sec:recompute-boundary}

Saving an expensive operator's output does not by itself bound the memory retained inside its exclude region. The operator may also save large inputs for backward, and many of those inputs are outputs of the upstream recompute region. Keeping them alive inside the save region defeats the intended release of upstream activations. For example, the profiled SFA query has shape $16{,}384\times32\times512$ in BF16 and occupies 512\,MiB; its positional-query input adds another 64\,MiB. The DSA indexer's query projection has shape $16{,}384\times64\times128$ in BF16 and occupies 256\,MiB. Retaining such inputs across many layers can consume substantial memory even when only a few operators are excluded.

\begin{figure}[htbp]
  \centering
  \includegraphics[width=\linewidth]{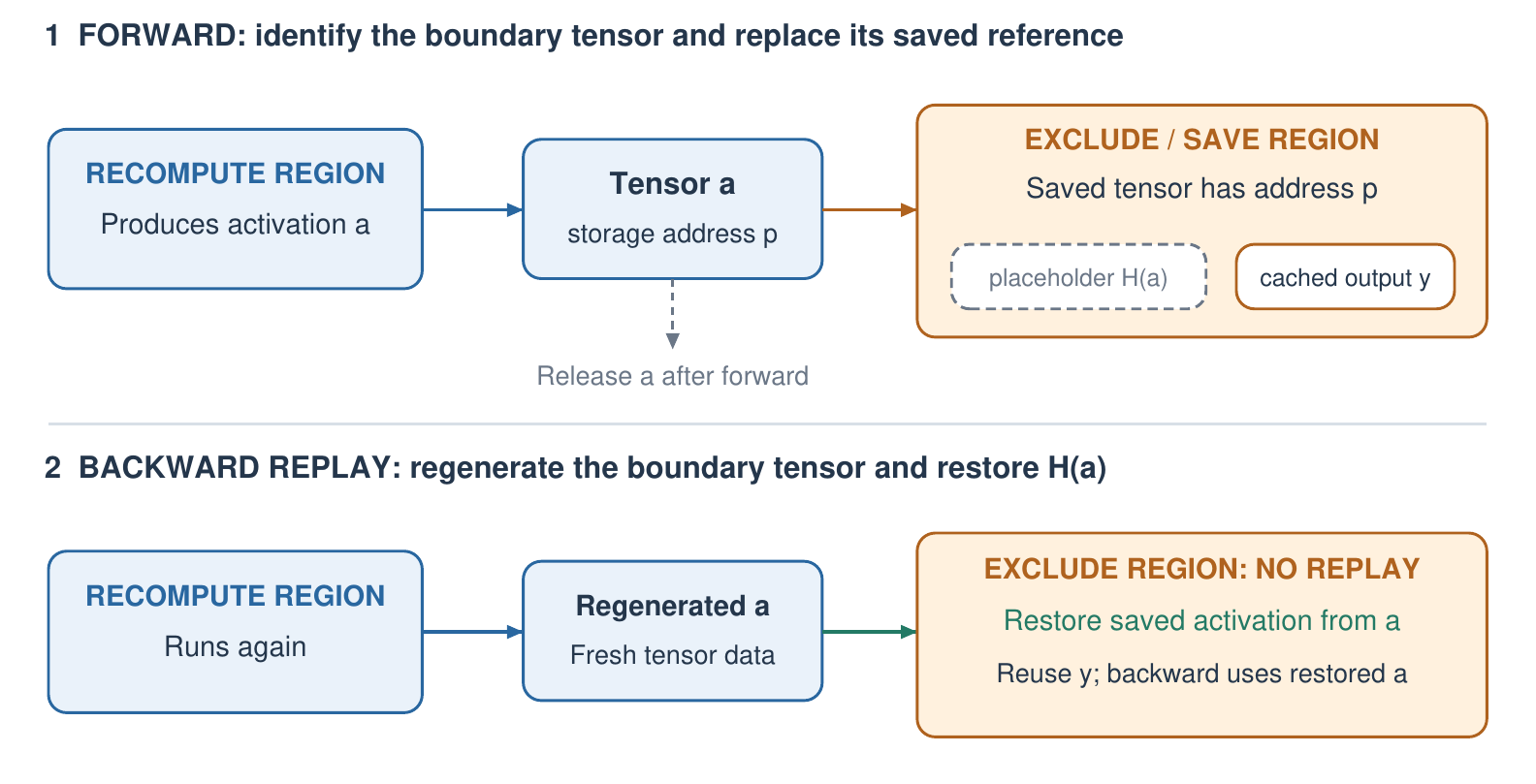}
  \caption{Boundary-activation memory optimization. Forward identifies a tensor crossing from recompute to save and replaces its saved reference with a placeholder when the storage address matches. Backward replay regenerates the boundary tensor and restores the save region's activation; the excluded operator reuses its cached output.}
  \label{fig:recompute-memory}
\end{figure}

We optimize the recompute-to-save boundary by recognizing the storage addresses of the activations crossing it. When the save region attempts to retain a tensor with a matching address, we store a placeholder instead of keeping the activation resident. Activations produced within the exclude region and the outputs needed to skip its execution are still saved. When backward triggers upstream recomputation, the newly generated boundary tensors restore the corresponding activations inside the save region. Figure~\ref{fig:recompute-memory} shows this process: the large boundary activation is released after forward and restored when needed, while the expensive excluded operator is not executed again. This removes the persistent boundary-input storage without adding another recomputation pass.

\subsection{DSA Long-Sequence Training Optimizations}
\label{sec:dsa}

DeepSeek Sparse Attention (DSA) uses a trainable indexer to select Top-K keys before performing sparse attention. For packed TND long-sequence training, an equal number of tokens does not imply an equal amount of computation: documents excluded by the mask may still occupy physical KV inputs, and auxiliary indexer graphs may remain live for a long time. We optimize the MindFormers/HyperParallel implementation along three boundaries: gradient scheduling, Q-block pairing, and the KV window, which structure the three subsections below.

\subsubsection{Gradient Scheduling: Layer-Wise Indexer Backward}

During the forward pass, the fused KL operator computes the gradient with respect to the indexer projection output. An explicit backward propagates this gradient to the indexer parameters. The teacher and main-branch inputs are detached from the auxiliary loss. An independent indexer graph can be consumed after each layer. If layers share an indexer graph, the auxiliary gradients are accumulated over the group and propagated once. With main-branch recomputation enabled, the framework retains the inputs needed for early backward and excludes the indexer and its loss from replay. Top-K indices, backward inputs, and shared projections remain live until their consumers finish.

Gradients are still accumulated according to the original gradient-accumulation (GAS) and loss-scaling rules, and the optimizer still updates at the end of the step. FSDP, recomputation input saving, and cache release cooperate to keep the early backward correct.

We evaluated layer-wise indexer backward in a separate run from the selective recomputation experiment. Both used the 28.9 s HSDP + Muon baseline. Layer-wise backward reduced step time to 24.5 s (15.2\%, Table~\ref{tab:layerwise-perf}). Scheduling the indexer gradient earlier lets each auxiliary graph be retired as soon as its backward computation finishes.

\begin{table}[htbp]
  \centering
  \caption{Effect of layer-wise indexer backward, evaluated against the HSDP\,+\,Muon configuration (\(1{,}024\) NPUs, sequence length \(262{,}144\)).}
  \label{tab:layerwise-perf}
  \small
  \begin{tabular}{lccc}
    \toprule
    \textbf{Configuration} & \textbf{Step Time} & \textbf{Speedup} & \textbf{Step-time reduction} \\
    \midrule
    HSDP + Muon & \(28.9\,\mathrm{s}\) & \(1.00\times\) & --- \\
    HSDP + Muon + layer-wise indexer backward & \(24.5\,\mathrm{s}\) & \(1.18\times\) & \(15.2\%\) \\
    \bottomrule
  \end{tabular}
\end{table}

\begin{figure}[t]
  \centering
  \includegraphics[width=0.94\linewidth]{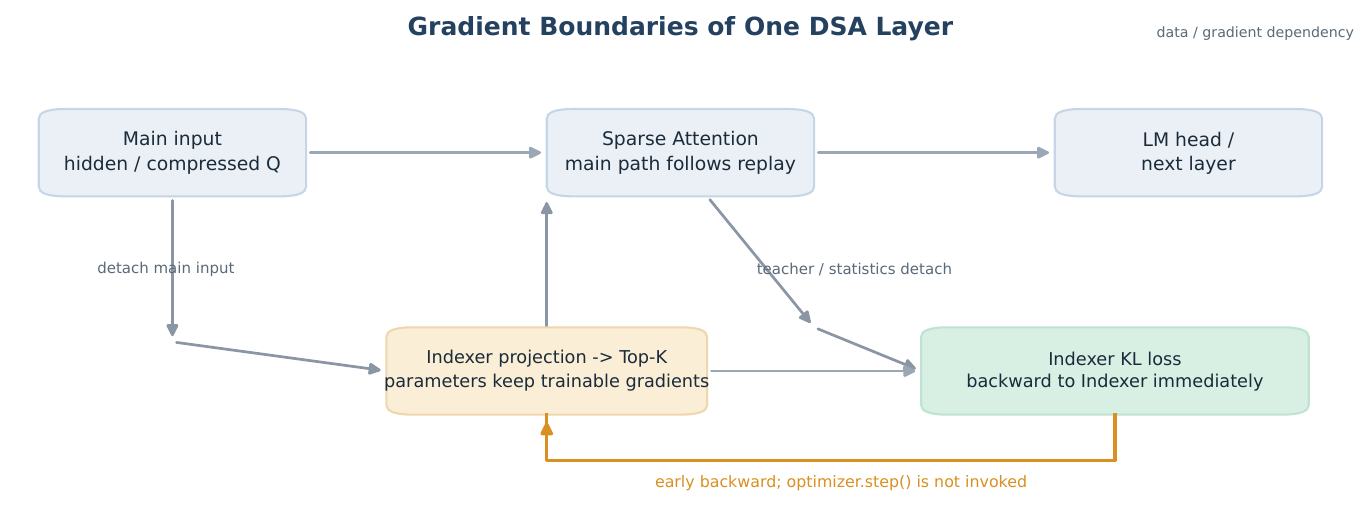}
  \caption{Indexer auxiliary-gradient boundaries: main-branch inputs and teacher statistics are isolated, the auxiliary KL loss is backpropagated early, and the optimizer still updates at the end of the step.}
  \label{fig:dsa-indexer}
\end{figure}

\subsubsection{Q-Block Pairing and Layout Planning}

Q-side load balancing is applied at three levels: within a single sequence, across the layers of one replica, and across data-parallel replicas.

Under a causal mask, later query positions can attend to more keys. At a fixed block size, head count, and feature dimension, the work of scanning candidate keys scales approximately with the number of visible query--key pairs. This applies to LightningIndexer before Top-K selection, dense flash attention, and the dense loss. Pairing an early block with a late block balances this component of the workload.

For sparse flash attention and its gradient (SFA/SFAG), Top-K selection bounds the number of keys processed by each query. Kernel time also depends on document boundaries, physical KV length, memory-access patterns, tiling, and gradient initialization and write-back. These effects motivate the use of measured operator costs in dynamic pairing.

\begin{figure}[t]
  \centering
  \includegraphics[width=0.94\linewidth]{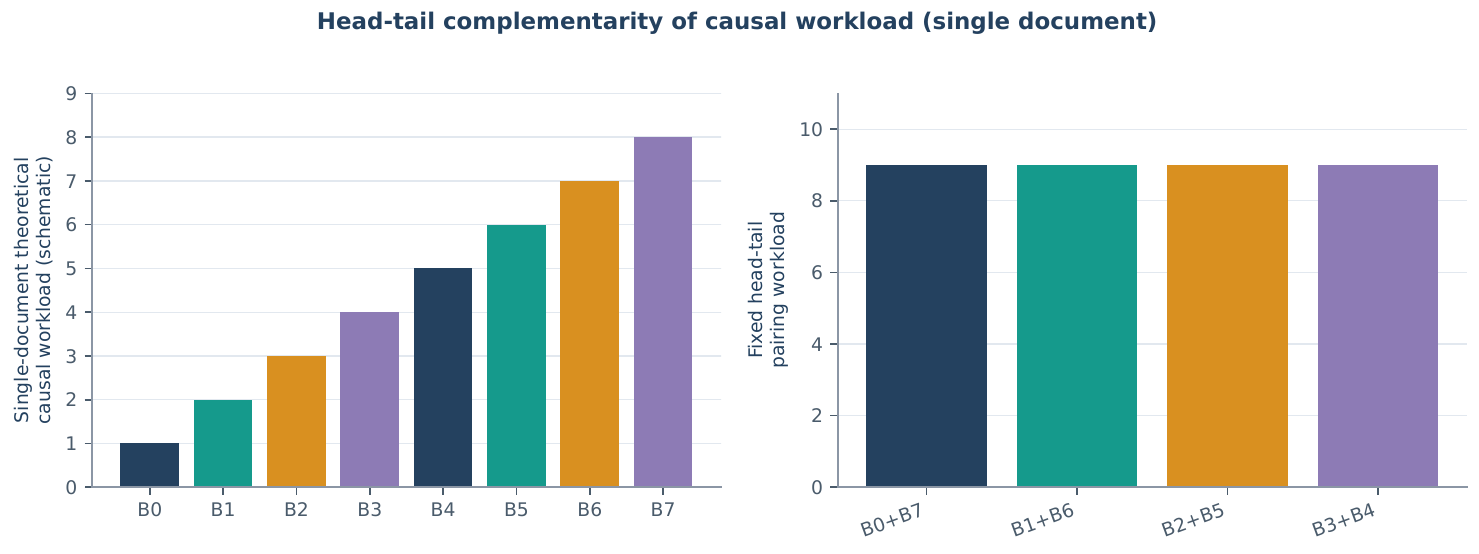}
  \caption{Head--tail complementarity of the theoretical causal workload for a single document. Values are abstract units and are not measured SFAG time.}
  \label{fig:dsa-causal}
\end{figure}

Figure~\ref{fig:dsa-causal} illustrates head--tail pairing with eight equal-length blocks whose causal workloads range from 1 to 8. With a contiguous split, the rank loads range from $1+2=3$ to $7+8=15$. Pairing blocks from opposite ends assigns a workload of 9 to each rank while preserving the total. These values count causal work in the example. The placement of the folding and unfolding operations determines the communication needed to realize the pairing.

For a packed multi-document sequence in TND layout, fixed folding partitions the sequence into $2R$ equal-length query blocks and assigns a head--tail pair to each of the $R$ ranks. Causal positions reset at every document boundary. A block near the end of the sequence can therefore contain queries with few visible keys. The residual imbalance depends on the lengths and positions of the packed documents.

\begin{figure}[t]
  \centering
  \includegraphics[width=0.80\linewidth]{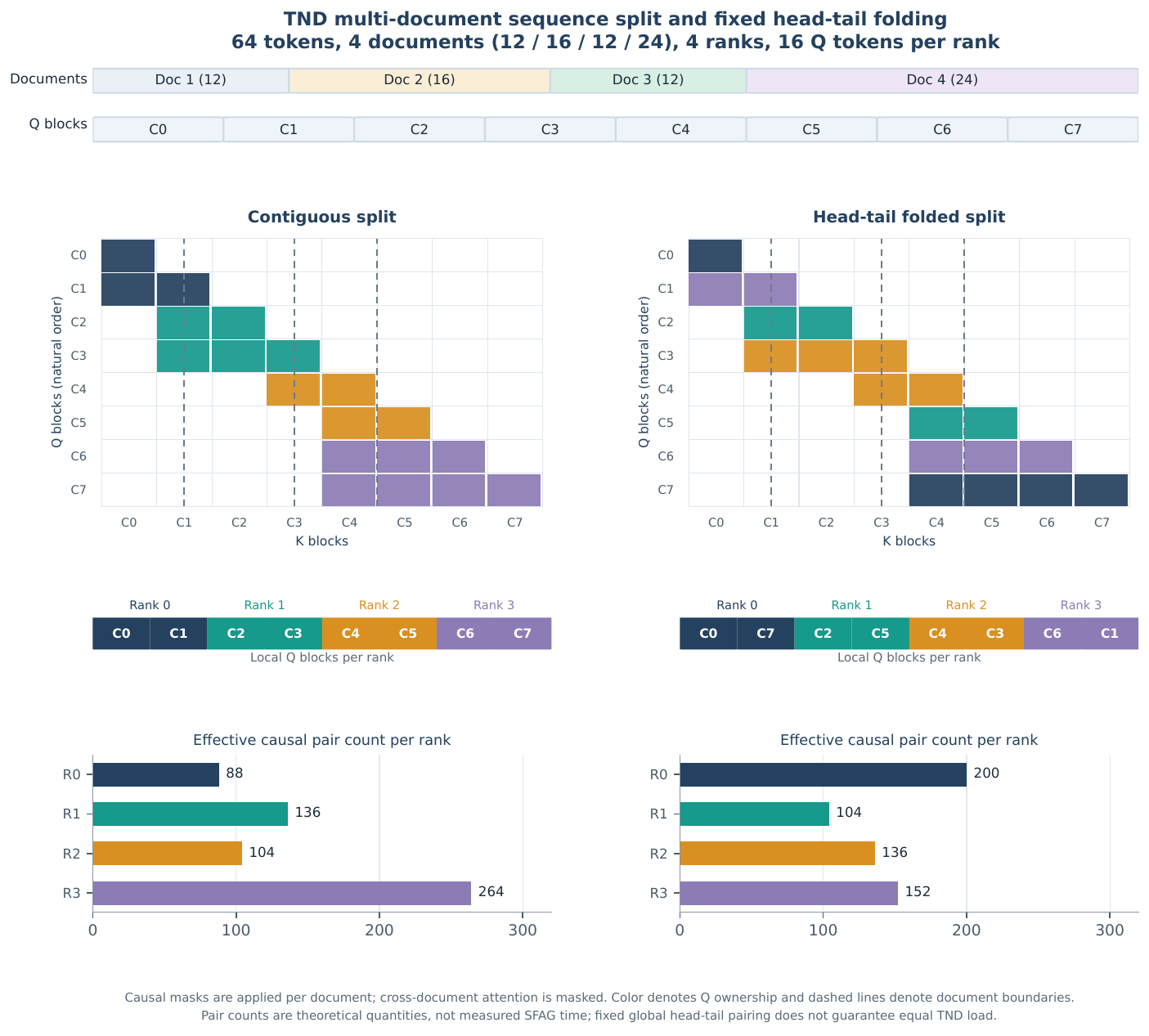}
  \caption{Fixed folding for TND multi-document sequences: causal positions reset at document boundaries, so head--tail complementarity no longer guarantees exactly equal load.}
  \label{fig:dsa-tnd}
\end{figure}

Dynamic pairing estimates cost on the same two-block layout using features such as in-document dense/sparse pair counts, logical KV, and active-document KV, and then pairs light with heavy blocks. It replaces the fixed block ownership only when the predicted peak decreases strictly; otherwise the fixed scheme is kept. Both schemes share the folding execution and gradient-restoration framework. In the network-level implementation of the sparse stage described here, the pairing table decides the selection of local input blocks at the model entry rather than renegotiating P2P communication at every operator or every layer. Inputs, labels, masks, and \texttt{position\_ids} are selected together, and forward, backward, and recomputation share one immutable plan; non-per-token objectives preserve the original normalization semantics through equivalent weights using the original owner denominators.

\begin{figure}[t]
  \centering
  \includegraphics[width=0.94\linewidth]{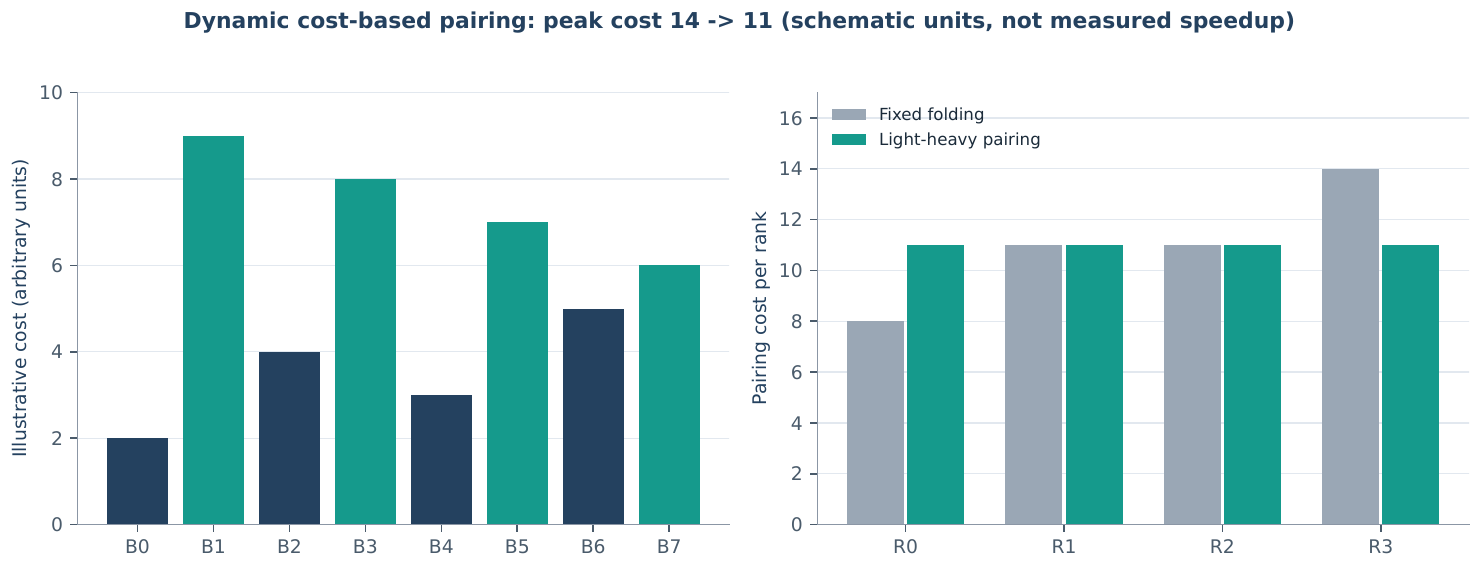}
  \caption{Illustration of dynamic cost-based pairing. $14 \rightarrow 11$ is an illustrative cost, not a measured speedup.}
  \label{fig:dsa-pairing}
\end{figure}

Once the pairing is decided, the folded layout must be applied consistently across the network.

The reference operator-boundary folding exchanges or redistributes Q-side tensors at the operator input and immediately unfolds the output after the operator finishes; if Indexer, SFA, and KL each return to the original layout, the same folding boundary is crossed repeatedly. In the sparse stage of this work, folding is planned at the model entry instead: assuming the complete input is already available, the corresponding blocks of \texttt{input\_ids}, \texttt{labels}, \texttt{loss\_mask}, and \texttt{position\_ids} are selected locally to form the folded token ownership before Embedding and the main-branch computation. Subsequent layers and all three DSA operators reuse this ownership; inside an operator the two Q blocks are only split locally and the required KV is selected, so Q tail slices are no longer exchanged per operator solely for folding and per-Q outputs are no longer moved back to their original rank.

We plan the CP and TP sequence layouts jointly. Each CP rank arranges its input so that the existing TP sequence distribution assigns complementary Q blocks to its TP ranks. The same layout is reused in backward and recomputation, avoiding a separate round trip for folding high-dimensional activations. KV AllGather, gradient reduction, TP layout conversions, and FSDP collectives still operate on their respective tensors. Positional encoding follows the original position IDs. MTP applies the shift within each folded block, preserving the supervision mask and loss normalization without a global redistribution of the final hidden states.

\begin{figure}[t]
  \centering
  \includegraphics[width=\linewidth]{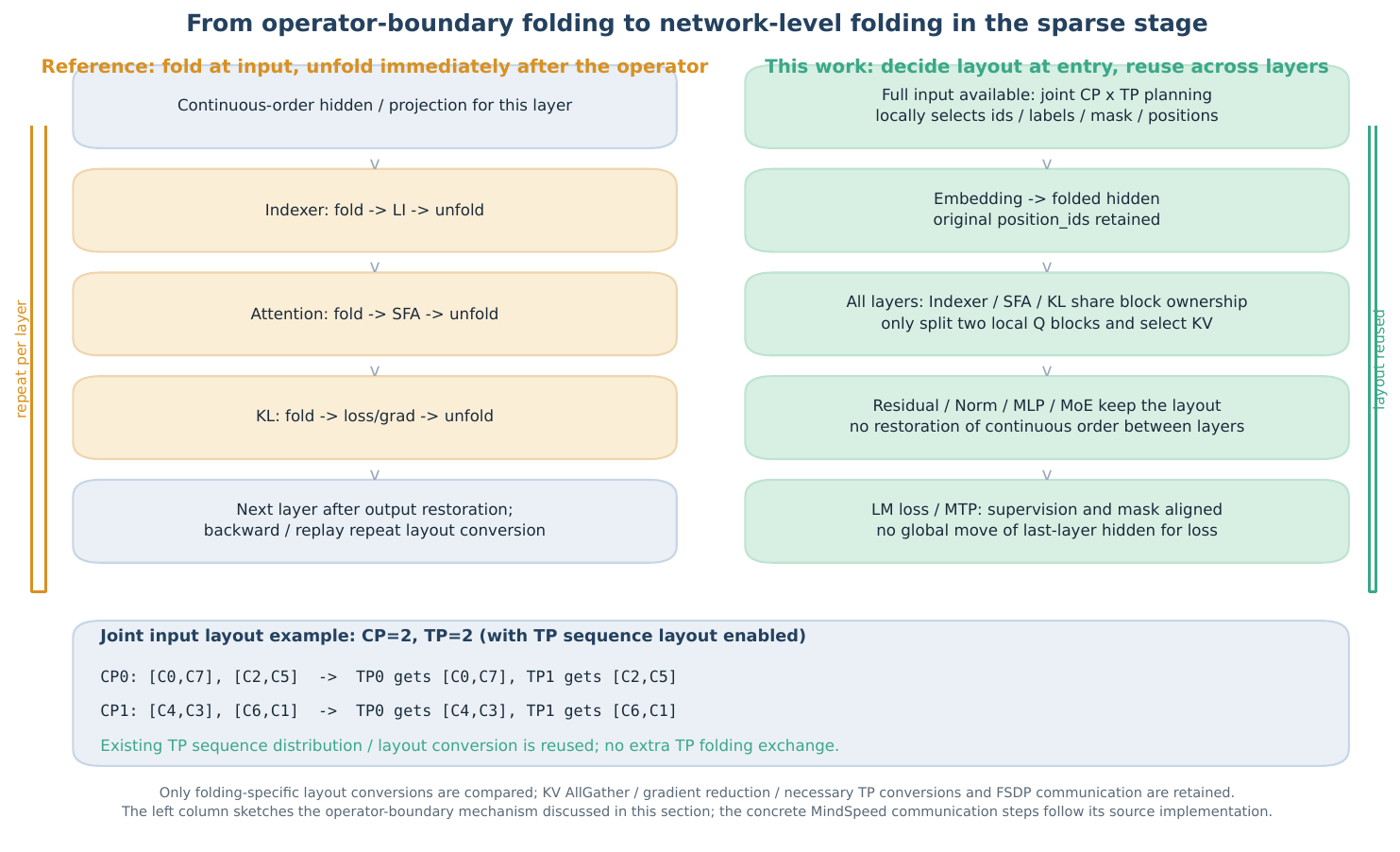}
  \caption{Operator-boundary immediate unfolding versus network-level folding in the sparse stage. Orange marks the extra boundary conversions of the reference path; green marks entry planning and folded-layout reuse. Existing KV/TP/FSDP communication is not removed by this change.}
  \label{fig:dsa-folding-scope}
\end{figure}

The network-level layout is used during sparse DSA training with synchronous CP and a single pipeline stage. During dense warm-up, the teacher attention consumes the main branch in its original token order, so the DSA operators fold their Q inputs within each layer. Figure~\ref{fig:dsa-folding-scope} locates the folding boundary in each scheme.

Folding and dynamic pairing within CP/TP can only adjust Q-block load inside one replica; they cannot remove the workload differences that arise when different DP replicas receive different document combinations. Under TND, even when every sample has the same token count after padding, document lengths, causal scan volumes, and active KV may differ. Inter-DP data balancing estimates cost before model execution from lightweight document-boundary metadata: it can use the dense proxy $\sum_i L_i^2$, or the DSA cost model to estimate the cost of the heaviest sequence shard in a complete TND micro-batch; for $LBS>1$, scoring uses the full layout with $T = LBS \times \text{sequence length}$.

The current step-aware implementation first follows the framework's native shuffle and data-fetch order to obtain the global sample window of this step, then treats each original micro-batch as an indivisible unit, stably sorts them by estimated cost, and assigns them to DP replicas and micro slots in a SNAKE pattern. A candidate is checked against both the maximum cumulative replica cost $\max_d \sum_m C[d,m]$ and the sum of per-micro-slot maxima $\sum_m \max_d C[d,m]$; it is adopted only when neither proxy gets worse and at least one improves. The global sample set, each DP quota, and the composition inside a micro-batch remain unchanged per step, so re-packing cannot change the local denominators of CE/CP/MTP; prefetching only advances a virtual plan, and the state is committed after successful consumption.

\begin{figure}[t]
  \centering
  \includegraphics[width=\linewidth]{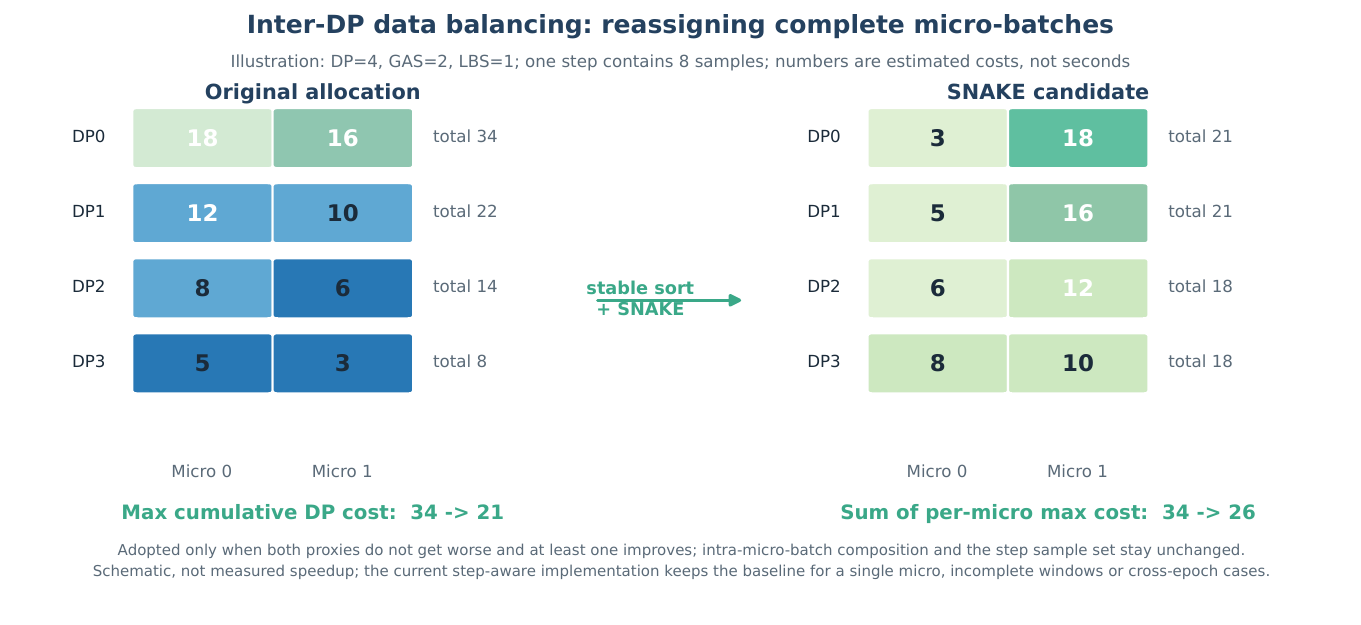}
  \caption{Scheduling illustration for DP=4 and GAS=2. The two cost proxies change from 34/34 to 21/26; the values are abstract estimated costs, not measured time or speedup.}
  \label{fig:dsa-dp-balance}
\end{figure}

The DP scheduler retains the original assignment when a replica has only one micro-batch, the scheduling window is incomplete, it crosses an epoch boundary, or the estimated cost fails the acceptance criterion. It assigns intact micro-batches across replicas and scores their cost under fixed folding. Within each replica, the CP/TP scheduler selects Q-block pairs using its own cost estimate. Document reordering within a micro-batch is a separate operation. The cost estimates in Figure~\ref{fig:dsa-dp-balance} describe the DP assignment alone.

\subsubsection{KV Window: Document-Aware KV Compact}

With a document-isolated causal mask, a Q block $B=[a,b)$ only needs the KV from the start $d(a)$ of its first active document up to $b$, i.e. $[d(a), b)$, rather than the original $[0,b)$ prefix. The implementation selects the intersecting chunks according to the physical arrangement of the current forward, trims boundaries, rewrites the Q/K cumulative lengths consistently, and removes empty segments and duplicated trailing padding. Indexer, SFA, and KL share the same coordinates. The backward sums gradients over overlapping intervals, then writes them back into the full KV and restores the saved arrangement so that the existing gradient-communication semantics are preserved.

\begin{figure}[t]
  \centering
  \includegraphics[width=0.94\linewidth]{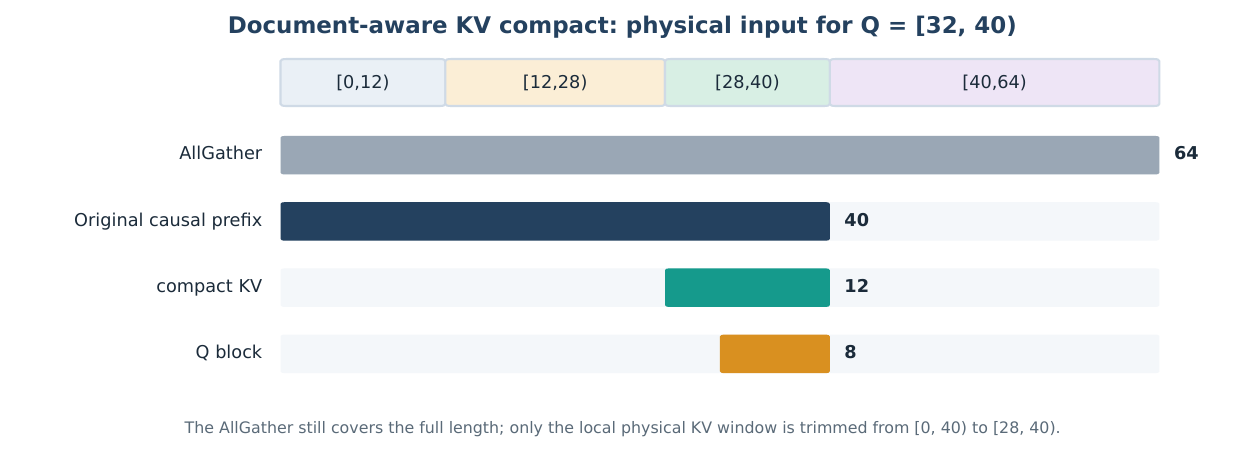}
  \caption{For $Q=[32,40)$, the physical KV is reduced from $[0,40)$ to $[28,40)$; the AllGather still covers the full length.}
  \label{fig:dsa-kv-compact}
\end{figure}

To validate the combined pipeline, companion operator regressions verify document boundaries, Top-K coordinates, overlapping-gradient write-back, and interleaved plans; a weighted combined comparison on 8 devices / 8 layers / 16K with TP2/CP4 ran 8 steps in each configuration, with maximum absolute differences of 0.000530 and 0.000441 for the main loss and MTP loss.

\section{Post-Training Framework}
\label{sec:post_train_optimization}

For reasoning and agentic capability development in Xing 4, we build a unified reinforcement-learning infrastructure. The framework connects task execution, experience collection, reward evaluation, and policy optimization in a common training loop. TeleGym provides the environment interface for heterogeneous task harnesses, while an intermediate server converts their interactions and outcomes into training-ready trajectories. This organization brings mathematical reasoning~\citep{zhao2025enhancing}, code generation~\citep{yang2025code}, and multi-turn agent~\citep{bianagentvocab,li2026memory} tasks into a shared training workflow while preserving task-specific execution and scoring semantics.

The system is organized around two requirements: sustaining training-data production across tasks with different execution lengths, and preserving the relationship between sampled actions, policy versions, and rewards as the model evolves. Disaggregated asynchronous execution combines partial rollout with oversampling to decouple trajectory completion from policy updates. Consistent trajectory records and execution-level reward processing connect this scheduling design to the learning objective. The following sections describe the overall training framework, TeleGym's execution architecture, and asynchronous rollout and policy optimization.

\subsection{Overall Training Framework}
\label{sec:rl-workflow}

\begin{figure}[htbp]
    \centering
    \includegraphics[width=\linewidth]{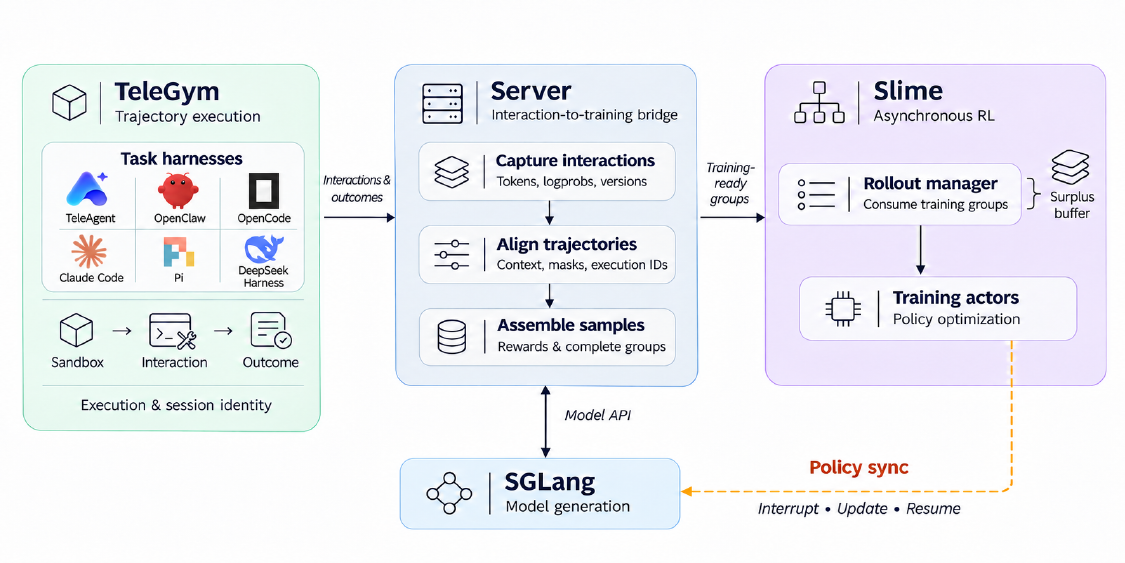}
    \caption{TeleGym executes agent tasks, the server converts interactions and outcomes into training groups, and Slime coordinates policy optimization. The dashed path denotes policy synchronization. Direct mathematical rollout is omitted for clarity.}
    \label{fig:rl-telegym-framework}
\end{figure}

Agentic training must accommodate harnesses with different tool interfaces, software dependencies, and interaction patterns. We organize the framework into three cooperating layers, as illustrated in Figure~\ref{fig:rl-telegym-framework}. TeleGym provides a shared execution layer for these harnesses, managing task environments and agent sessions. An intermediate server connects their model requests to the policy service and converts the resulting interactions into training trajectories. Slime coordinates rollout scheduling and policy optimization. Together, these layers connect heterogeneous agent executions to a common reinforcement-learning loop.

For each task, the dispatcher supplies the prompt, workspace inputs, and execution configuration to TeleGym. The selected harness interacts with its environment and accesses SGLang generation through the server's model API. Code tasks are evaluated through executable tests; general agent tasks support automatic, model-based, or hybrid evaluation of transcripts and workspace artifacts. Mathematical tasks use direct rollout and answer grading.

The server records sampled tokens, behavior-policy log probabilities, and policy versions, and associates these records with task outcomes. It then aligns context and response tokens, constructs generated-token masks, and assembles scored executions into training groups. Request and execution identities connect model interactions to asynchronously returned scores. These groups enter a shared policy-optimization pipeline, allowing task execution and evaluation to vary without changing the trainer's input representation.

For multi-turn tasks, one execution may produce several training segments but remains a single independent attempt. Group-relative reward processing therefore operates over executions of the same prompt before propagating the resulting training signal to their constituent segments. This prevents tasks with more model calls from being counted as additional attempts in group statistics. Only acknowledged task outcomes enter group assembly, with missing scores or evaluation failures distinguished from valid unsuccessful attempts.

\subsection{TeleGym: Unified Agent Execution}
\label{sec:rl-telegym}

\begin{figure}[htbp]
    \centering
    \includegraphics[width=\linewidth]{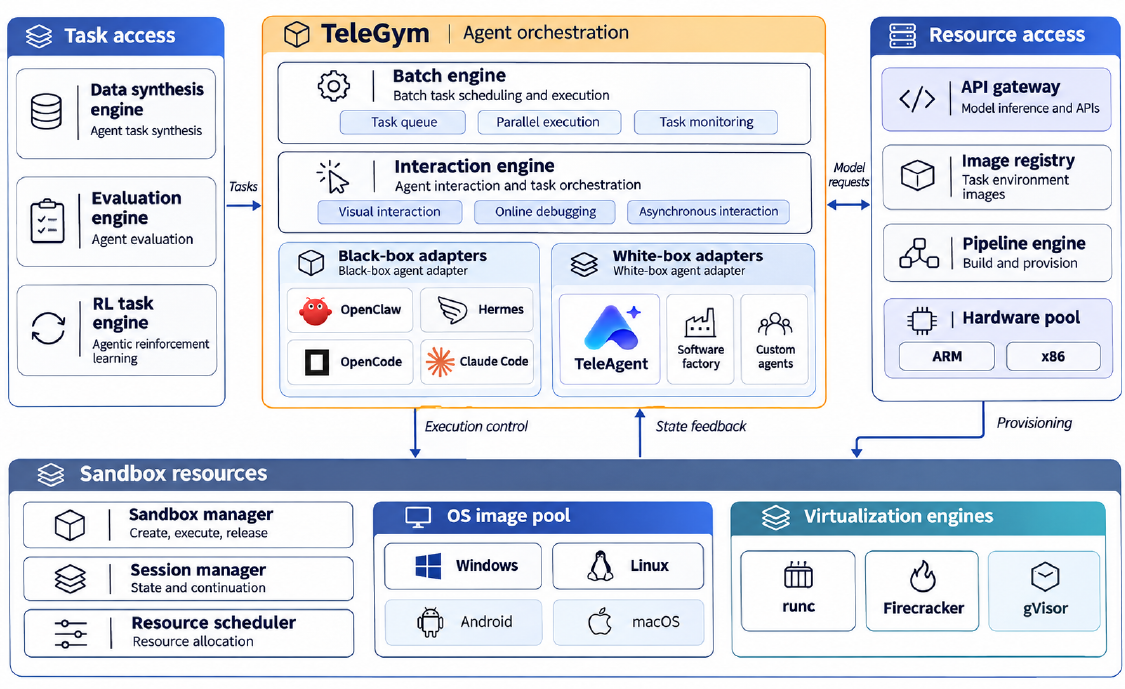}
    \caption{TeleGym platform architecture. Task engines connect to batch and interaction engines through the orchestration layer, which accommodates black-box and white-box agent adapters. Resource access connects model APIs, environment images, and deployment services to the sandbox layer. OS images, virtualization engines, and hardware pools describe the platform's integration scope; the RL execution path discussed here uses the common runtime and sandbox interfaces.}
    \label{fig:telegym-architecture}
\end{figure}

Training across heterogeneous agents requires a common way to prepare task environments, run different harnesses, and maintain interactions over multiple turns. We use TeleGym as the execution and orchestration layer between task engines and the underlying model and sandbox services. Figure~\ref{fig:telegym-architecture} presents the broader platform architecture through task access, agent orchestration, resource access, and sandbox management. Within this architecture, the RL execution path turns a dispatched task into a managed agent episode, connecting its workspace, harness session, and model interactions throughout execution. This provides a reusable execution substrate for data synthesis, evaluation, and reinforcement learning~\citep{zou2026group, lv2026beyond}. We describe four aspects relevant to this training workflow below.

\paragraph{Task-facing execution.}
Task engines provide the problem specification, workspace inputs, and runtime configuration, while retaining control over task selection and evaluation. The platform distinguishes batch orchestration, organized around task queues, parallel execution, and monitoring, from interactive orchestration, which provides visual interaction, online debugging, and asynchronous interaction. At the runtime boundary, both use a common episode lifecycle for initialization, execution, continuation, and cleanup. Batch workflows invoke this interface across independent tasks; interactive workflows advance an existing episode through successive actions. This allows task-specific synthesis, evaluation, or RL logic to reuse environment preparation and execution mechanisms without duplicating harness integration.

\paragraph{Harness-independent orchestration.}
TeleGym exposes a common episode interface while allowing each harness to retain its own agent loop and tool semantics. The platform distinguishes black-box integration of external harnesses from white-box integration of native or custom agents, including TeleAgent. Runtime adapters translate task actions into harness-specific invocations and normalize execution results. This accommodates different agents without requiring the task engine to implement their individual invocation and session protocols. An action may include several model calls and tool interactions, so the interface supports extended agent workflows as well as short executions. Harness adaptation and sandbox adaptation form separate interfaces, allowing either component to be extended without changing the task-facing execution contract.

\paragraph{Sandbox and session management.}
Sandbox adapters separate agent execution from deployment on local, containerized, or remote resources. For each active episode, TeleGym binds a runtime to a sandbox context, prepares workspace files and dependencies, and initializes the required tools. Image and deployment services supply the underlying execution resources, while the adapter exposes common command-execution and file-access operations. TeleGym retains the harness session across successive actions so that multi-turn execution proceeds in the same task context. Workspace artifacts remain accessible for evaluation before environment cleanup. Episode identity connects these operations and artifacts to the originating task, preserving a coherent execution context across model calls and intervening tool use.

\paragraph{Integration with the training loop.}
TeleGym configures each runtime to access the policy through the external model gateway and propagates episode-level association metadata. This connects environment-side execution to the model-side records collected by the server: TeleGym manages the agent session and workspace, while the server constructs the token-level trajectories consumed by the trainer. Keeping these responsibilities separate allows the training pipeline to support different harnesses through a shared data interface. Session continuity is maintained at the episode level; policy publication and interruption of active generation are coordinated by the rollout controller, as described next.

\subsection{Task Synthesis}
\label{sec:rl-data-synthesis}

Data synthesis prepares executable tasks for general and code agents, each combining an instruction, an initial environment, and evaluation criteria. Multiple harnesses participate in both task design and selection: agents operating through different harnesses propose and refine candidate tasks, which are then independently attempted across harnesses. TeleGym provides a common runtime for these executions, linking design feedback and validation outcomes to each candidate task.

\paragraph{General agents.}
General-agent tasks are constructed around representative workflows involving information retrieval, document processing, and tool use. Instructions are paired with relevant files and tool interfaces, with variations in constraints and interaction steps to cover both single-turn and multi-turn scenarios. Feedback from multiple harnesses guides refinement of task instructions, tool requirements, and evaluation criteria. Validation runs start from equivalent task conditions and assess outputs or environment state against shared criteria, using programmatic checks or model-based assessment as appropriate.

\paragraph{Code agents.}
Code-agent tasks are constructed around software projects and programming requirements, including implementation and bug fixing. Agents using multiple harnesses contribute to task formulation and review of the associated tests. Candidate tasks are then attempted independently from the same initial code workspace and dependency configuration. Shared executable tests verify both the requested changes and preservation of existing functionality. These runs expose environment issues and mismatches between task descriptions and tests before selection for training.

Across both categories, task validation and deduplication remove unusable or repetitive examples. Accepted tasks enter the shared rollout pipeline through TeleGym's execution interface.

\subsection{Asynchronous Training with Partial Rollout}
\label{sec:rl-async}

Agentic rollouts vary in both generation length and environment execution time. Tool calls, workspace operations, and reward evaluation can extend a task beyond the completion of other samples dispatched with it. We combine disaggregated asynchronous execution, bounded oversampling, and partial rollout to accommodate this variability. The design separates three boundaries: completion of a task, assembly of a training batch, and publication of an updated policy. A long-running task can therefore span multiple policy updates without requiring all concurrent tasks to finish together.

\paragraph{Disaggregated execution.}
Generation and policy optimization run on separate device pools, with the inference service remaining resident while the trainer updates the policy. The rollout pipeline can prepare subsequent training data during optimization of the current batch. TeleGym manages environment-side execution, the server associates interactions with task outcomes, and the rollout manager supplies eligible groups to the trainer. Optimization starts once enough complete, scored groups are available. This producer--consumer organization allows model generation, tool execution, and reward processing to progress independently of individual optimizer steps. Policy publication remains an explicit coordination point between the training and inference pools.

\begin{figure}[htbp]
    \centering
    \includegraphics[width=\linewidth]{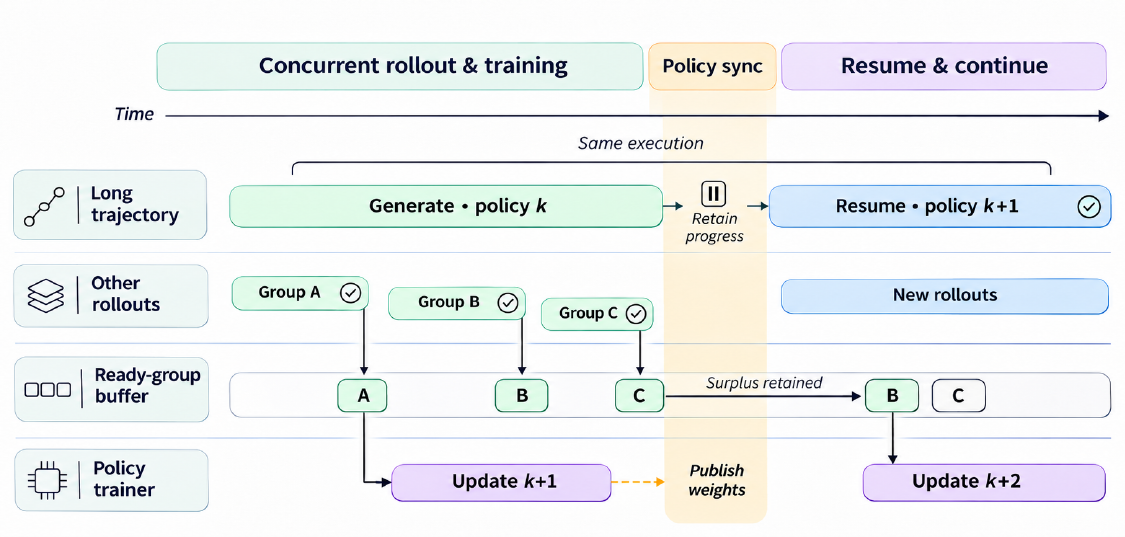}
    \caption{A long trajectory pauses at policy synchronization and resumes under the updated policy. Group A supports an earlier update; surplus groups B and C remain buffered, with B consumed later. Counts and durations are schematic.}
    \label{fig:rl-partial-rollout}
\end{figure}

\paragraph{Bounded oversampling.}
The scheduler authorizes more rollout work than the next training batch requires, providing additional candidates when some tasks finish late or their groups are removed by reward-based filtering. A finite lookahead window limits how far task production can advance beyond training consumption; separate bounds on pending and ready groups constrain unfinished work and accumulated training data. Together, these limits prevent a fast producer or a slow consumer from creating an unbounded backlog. Completed executions are assembled into query groups, and only eligible groups enter the ready buffer. Groups in excess of the current batch remain available for later updates, as illustrated in Figure~\ref{fig:rl-partial-rollout}. Retention carries unconsumed data forward; it does not by itself prescribe repeated training on previously consumed groups.

\paragraph{Policy publication and trajectory continuation.}
The policy-publication interval is configurable and need not coincide with every rollout completion. At publication, the controller pauses new model-request admission, interrupts active generation, and waits for model execution to reach a safe synchronization boundary. It then installs the updated weights, advances the policy version, and reopens admission. Weight replacement cannot proceed while old model execution remains active. This ordering prevents an uncontrolled weight change within an executing generation segment while allowing the enclosing task to continue across policy versions.

Unfinished trajectories retain their execution identity and progress and resume after synchronization. For a multi-turn agent, this preserves the association between earlier responses, tool interactions, and subsequent model calls under the updated policy. The synchronization boundary concerns model execution rather than completion of the entire agent session, which may still require further environment interaction or evaluation. Partial rollout preserves work in progress: interruption alone neither assigns a terminal reward nor makes an unfinished prefix eligible for training.

\paragraph{Policy lag and sample eligibility.}
Resumed trajectories and buffered groups may contain samples generated by earlier policy versions. Each generated segment retains its sampling-time log probabilities, which provide the behavior-policy probabilities used in importance-ratio computation during optimization. Recomputing probabilities with updated weights does not replace this record. Optional stale-data filtering uses observed policy versions to identify old trajectories and removes the entire query group when a constituent trajectory is rejected, preserving the membership used for group-relative reward processing. This is a trajectory-level heuristic rather than a strict token-wise age bound. These mechanisms address policy lag separately from the scheduling limits that bound outstanding work.

\paragraph{Checkpoint boundaries.}
Checkpointing additionally coordinates the state of data production and consumption. At a save point, the scheduler stops extending the authorized task window and completes the outstanding window before persisting unconsumed groups, data-source progress, execution identities, and version metadata. Model and optimizer state are saved after this data state has been recorded. This differs from routine policy publication, which interrupts and resumes unfinished trajectories. On restart, the saved buffer and cursor preserve the outstanding data stream, while a new execution-generation namespace separates new work from callbacks associated with the previous run.

\section{Training Cluster Performance Optimization}
\label{sec:cluster_optimization}

\begingroup
\raggedbottom

Host scheduling is a further source of overhead in large-scale training. In the Xing 4 training infrastructure, host CPUs prepare data, submit device tasks, and coordinate communication, while Ascend accelerators perform most of the numerical computation. Contention for host CPU time delays the execution of ready training threads. When these delays extend beyond the overlap with other training work, they leave accelerators waiting for work or prolong collective completion on the training step's critical path. Host resource management therefore complements the device-side optimizations described in Section~\ref{sec:training_optimization}.

We combine CPU affinity with core isolation to limit host-side interference. Fine-grained affinity assigns training threads to CPU subsets according to their execution roles; core isolation separates selected background services and movable interrupt handlers from the training CPU pool. Figure~\ref{fig:cluster-contention} summarizes host-side sources of training delays and the scope of the two optimizations.

\begin{figure}[htbp]
    \centering
    \footnotesize
    \setlength{\tabcolsep}{4pt}
    \renewcommand{\arraystretch}{1.18}
    \begin{tabularx}{\linewidth}{@{}>{\raggedright\arraybackslash}X>{\raggedright\arraybackslash}X>{\raggedright\arraybackslash}X@{}}
    \toprule
    \rowcolor{black!6} \textbf{Host contention} & \textbf{Delayed execution} & \textbf{Training impact} \\
    Training threads compete with other threads and background work for CPU time. & Data preparation, task submission, or collective participation is delayed. & Delays exposed on the critical path increase device idle time or collective completion time. \\
    \midrule
    \textbf{Fine-grained affinity} & \multicolumn{2}{p{.63\linewidth}@{}}{Training threads execute within CPU subsets assigned by execution role.} \\
    \textbf{Core isolation} & \multicolumn{2}{p{.63\linewidth}@{}}{Selected background services and movable interrupt handlers execute outside the training CPU pool.} \\
    \bottomrule
    \end{tabularx}
    \caption{Host CPU contention, training delays, and the scope of fine-grained affinity and core isolation.}
    \label{fig:cluster-contention}
\end{figure}

\subsection{Fine-Grained CPU Affinity and Core Isolation}
\label{sec:cluster-affinity}

On each host, let \(T\) denote the set of logical CPUs allocated to training. For a configuration with one training process per device, \(T_i\subseteq T\) denotes the pool assigned to local device \(i\). Fine-grained affinity partitions \(T_i\) into disjoint subsets by execution role. MindSpore supports thread-level affinity for its main, runtime, PyNative, and MindData threads~\citep{mindspore-msrun,linux-sched-setaffinity}. The allocation described here uses three groups: PyNative execution; MindData loading and preprocessing; and the main thread together with runtime and training-support threads. Separate subsets limit scheduling competition between these groups. Subset sizing follows the CPU demand of each group, balancing data preparation, execution, and runtime support within the per-device pool.

Core isolation addresses interference from outside the training process. On each host, a separate CPU pool \(H\), disjoint from \(T\), accommodates selected background services, logging, monitoring, and movable interrupt handlers. Task placement controls assign the selected services to \(H\), while IRQ affinity controls direct movable interrupts to that pool. The isolation policy targets these configurable sources of interference, reducing competition for training CPUs while reserving capacity for system work~\citep{linux-cpu-isolation}. Figure~\ref{fig:cluster-placement} illustrates the separation between role-specific training subsets and the system pool. CPU identifiers and subset sizes are determined by host topology and workload requirements.

\begin{figure}[htbp]
    \centering
    \footnotesize
    \setlength{\tabcolsep}{4pt}
    \renewcommand{\arraystretch}{1.18}
    \begin{tabularx}{\linewidth}{@{}>{\raggedright\arraybackslash}X>{\raggedright\arraybackslash}X@{}}
    \toprule
    \rowcolor{black!6} \textbf{Training pool: role subsets of \(T_i\)} & \textbf{System pool: \(H\cap T=\emptyset\)} \\
    PyNative execution threads & Selected background services \\
    MindData loading and preprocessing & Logging and monitoring processes \\
    Main, runtime, and training-support threads & Movable interrupt handlers \\
    \midrule
    Separate CPU subsets limit competition between training roles. & A separate CPU pool limits interference from selected system work. \\
    \bottomrule
    \end{tabularx}
    \caption{CPU allocation by execution role: thread categories within the per-device training pool and the separate system pool.}
    \label{fig:cluster-placement}
\end{figure}

Linux CPU affinity is a per-thread property. A thread executes on online CPUs permitted by both its affinity mask and its effective cgroup cpuset. The cpuset controller also constrains the memory nodes available to its tasks. Host resource allocation accounts for NUMA topology and the relationship between logical CPUs and physical cores: on systems with simultaneous multithreading enabled, sibling logical CPUs share a physical core's execution resources. CPU affinity governs execution placement, while NUMA memory policy governs memory placement within the allowed nodes~\citep{linux-sched-setaffinity,linux-cgroup-v2}. These complementary controls define the CPU and memory boundaries of the host configuration.

\subsection{Training Throughput Results}
\label{sec:cluster-throughput}

The cluster study uses the 285B pretraining workload with MindSpore HyperParallel on Ascend 910C hardware. Initial validation used a 512-accelerator A3 cluster, followed by deployment on 8,192 accelerators. Table~\ref{tab:cluster-throughput} summarizes the relative throughput gains with CPU affinity and core isolation enabled together. At each cluster size, the gain is defined as \((Q_{\mathrm{opt}}/Q_{\mathrm{base}}-1)\times100\%\), where \(Q_{\mathrm{base}}\) and \(Q_{\mathrm{opt}}\) denote the pre-optimization and optimized training throughput, respectively. The combined configuration increased throughput by 1--2\% at 512 accelerators and by 5--10\% at 8,192 accelerators.

\begin{table}[htbp]
    \centering
    \caption{Training throughput gains with CPU affinity and core isolation enabled together, relative to the pre-optimization baseline at each cluster size.}
    \label{tab:cluster-throughput}
    \small
    \begin{tabularx}{\linewidth}{@{}>{\raggedright\arraybackslash}Xcc@{}}
    \toprule
    \textbf{Deployment} & \textbf{Ascend 910C accelerators} & \textbf{Relative throughput gain} \\
    \midrule
    CPU affinity + core isolation & 512 & 1--2\% \\
    CPU affinity + core isolation & 8,192 & 5--10\% \\
    \bottomrule
    \end{tabularx}
\end{table}

The combined configuration improves training throughput at both deployment scales. Fine-grained affinity organizes CPU resources within the training process, while core isolation separates selected system work from the training pool. Together, these host resource-management measures complement device-side optimization in the 285B pretraining deployment.

\endgroup

\section{Conclusion}
\label{sec:conclusion}

We have presented the architectural design principles and training infrastructure behind the Xing 4 series on Ascend. The open-sourced 29B model and the larger configuration studied here illustrate how sparse activation, attention, and expert width can be adapted to different parameter budgets. Simulations quantify the cost of long-context attention, and training measurements show how network depth affects throughput.

The pretraining system brings topology-aware hybrid sharding and owner-based Muon execution together with coordinated management of long-sequence workloads and activation memory. CPU affinity and core isolation address host scheduling interference that can propagate through distributed execution. Implemented with MindSpore HyperParallel, these mechanisms align model-state placement, matrix computation, communication, and host resources with the Ascend platform. The large-scale pretraining run documented in this report demonstrates their integration on 8,192 Ascend 910C cards.

We have introduced TeleGym, a unified agent execution platform, and integrated it into the reinforcement-learning infrastructure for the Xing 4 series. Shared execution interfaces and trajectory construction connect task synthesis, environment interaction, and reward evaluation to policy optimization. Disaggregated asynchronous execution, bounded oversampling, and partial rollout accommodate heterogeneous task durations and allow trajectories to continue across policy updates. Sampling-time policy records and execution-level reward grouping retain the correspondence between generated actions and learning signals, supporting a common training workflow for reasoning, coding, and multi-turn agents.

We are now training trillion-parameter models in the Xing 4 series on Ascend. These models use cmHC, which applies geometric contraction to regulate mixing across residual streams~\citep{he2026cmhc}. At this scale, we continue to study how longer contexts and more complex agent workloads affect training efficiency.

\bibliography{iclr2024_conference}

@misc{telechat,
      title={TeleChat Technical Report}, 
      author={Zhongjiang He and Zihan Wang and Xinzhang Liu and Shixuan Liu and Yitong Yao and Yuyao Huang and Xuelong Li and Yongxiang Li and Zhonghao Che and Zhaoxi Zhang and Yan Wang and Xin Wang and Luwen Pu and Huinan Xu and Ruiyu Fang and Yu Zhao and Jie Zhang and Xiaomeng Huang and Zhilong Lu and Jiaxin Peng and Wenjun Zheng and Shiquan Wang and Bingkai Yang and Xuewei he and Zhuoru Jiang and Qiyi Xie and Yanhan Zhang and Zhongqiu Li and Lingling Shi and Weiwei Fu and Yin Zhang and Zilu Huang and Sishi Xiong and Yuxiang Zhang and Chao Wang and Shuangyong Song},
      year={2024},
      eprint={2401.03804},
      archivePrefix={arXiv},
      primaryClass={cs.CL},
      url={https://arxiv.org/abs/2401.03804},
      note = {arXiv preprint arXiv:2401.03804}
}

@inproceedings{zou2026group,
  title={Group Structural Proximity Shapes Exploration-Exploitation in Tool-Use Reinforcement Learning},
  author={Zou, Lexiao and Yao, Yitong and Chen, Gongwei and Deng, Wenmin and Wang, Chao and Song, Shuangyong and Zhang, Miao},
  booktitle={Findings of the Association for Computational Linguistics: EMNLP 2026},
  pages={4682--4698},
  year={2026}
}

@inproceedings{lv2026beyond,
  title={Beyond SFT-Optimality: Cold-Start Data Selection for Reinforcement Learning via Gradient Subspace Coverage},
  author={Lv, Haisheng and Su, Zikun and Liu, Jingbang and Huang, Zilu and Liu, Xinzhang and Deng, Wenmin and Zhao, Yu and Wang, Chao and Song, Shuangyong},
  booktitle={Proceedings of the 2026 Conference on Empirical Methods in Natural Language Processing},
  pages={11769--11789},
  year={2026}
}

@article{zhao2025enhancing,
  title={Enhancing math reasoning ability of large language models via computation logic graphs},
  author={Zhao, Deji and Han, Donghong and Wu, Jia and He, Zhongjiang and Ning, Bo and Yuan, Ye and Li, Yongxiang and Wang, Chao and Song, Shuangyong},
  journal={Knowledge-Based Systems},
  volume={325},
  pages={113905},
  year={2025},
  publisher={Elsevier}
}

@article{zhang2024lemur,
  title={Lemur: Log parsing with entropy sampling and chain-of-thought merging},
  author={Zhang, Wei and Guan, Xiangyuan and Yunhong, Lu and Zhang, Jie and Song, Shuangyong and Cheng, Xianfu and Wu, Zhenhe and Li, Zhoujun},
  journal={arXiv preprint arXiv:2402.18205},
  year={2024}
}

@article{he2026cmhc,
  title={{cmHC}: Balancing Stream Mixing in Hyper-Connections via Geometric Contraction},
  author={He, Haowei and Zhao, Ziwen and Pan, Chenshuo and Lin, Gang and Yang, Zhihao and Wang, Chao and Liu, Xinzhang and Yu, Kaidong and Gao, Han and Li, Dongfang and Yang, Menglin and Song, Shuangyong and Li, Yongxiang},
  journal={arXiv preprint arXiv:2611.27193},
  year={2026},
  url={https://arxiv.org/abs/2611.27193}
}

@inproceedings{hu2026awakening,
  title={Awakening Dormant Experts: Counterfactual Routing to Mitigate MoE Hallucinations},
  author={Hu, Wentao and Zhai, Yanbo and Hu, Xiaohui and Zhao, Mingkuan and Liu, Xue and Yu, Kaidong and Song, Shuangyong and Li, Xuelong and others},
  booktitle={Proceedings of the 64th Annual Meeting of the Association for Computational Linguistics (Volume 1: Long Papers)},
  pages={47257--47271},
  year={2026}
}

@inproceedings{hu2026mosaic,
  title={Mosaic pruning: A hierarchical framework for generalizable pruning of mixture-of-experts models},
  author={Hu, Wentao and Zhao, Mingkuan and Song, Shuangyong and Zhu, Xiaoyan and Lai, Xin and Wang, Jiayin},
  booktitle={Proceedings of the AAAI Conference on Artificial Intelligence},
  volume={40},
  number={26},
  pages={21885--21893},
  year={2026}
}

@article{xiong2025tablezoomer,
  title={Tablezoomer: a collaborative agent framework for large-scale table question answering},
  author={Xiong, Sishi and He, Ziyang and He, Zhongjiang and Zhao, Yu and Pan, Changzai and Zhang, Jie and Song, Shuangyong and Li, Yongxiang},
  journal={Vicinagearth},
  volume={2},
  number={1},
  pages={11},
  year={2025},
  publisher={Springer}
}

@inproceedings{bianagentvocab,
  title={AgentVocab: Structure-Aware Vocabulary Adaptation for Efficient LLM Agents},
  author={Bian, Kai and Mo, Haosi and Liu, Xuebo and Song, Shuangyong and Li, Jing and Li, Yongxiang and Zhang, Min and Li, Xuelong},
  booktitle={Forty-third International Conference on Machine Learning}
  ,
  year={2026}
}

@article{li2026memory,
  title={Memory-Centric Continual Learning for AI Agents},
  author={Li, Dongfang and Chen, Zhuoen and Jia, Bonian and Liu, Zixuan and Lu, Zhengxuan and Deng, Can and Shi, Yukun and Meng, Yuhang and Zhang, Xin and Huang, Jiahe and others},
  journal={arXiv preprint arXiv:2610.29357},
  year={2026}
}

@article{yang2025code,
title={From code foundation models to agents and applications: A practical guide to code intelligence},
author={Yang, Jian and Zhang, Wei and Liu, Shark and Wu, Jiajun and Guo, Shawn and Li, Yizhi and Pan, Changzai and Zhang, Jie and Song, Shuangyong and etc.},
journal={arXiv preprint arXiv:2511.18538},
year={2025}
}

@inproceedings{wang2024telechat,
  title={TeleChat: An Open-source Billingual Large Language Model},
  author={Wang, Zihan and Yao, Yitong and Mengxiang, Li and He, Zhongjiang and Wang, Chao and Song, Shuangyong and others},
  booktitle={Proceedings of the 10th SIGHAN Workshop on Chinese Language Processing (SIGHAN-10)},
  pages={10--20},
  year={2024}
}

@misc{wang2025telechat2,
  title = {Technical Report of TeleChat2, TeleChat2.5 and T1},
  author = {Wang, Zihan and Liu, Xinzhang and Yao, Yitong and Wang, Chao and Zhao, Yu and Yang, Zhihao and Deng, Wenmin and Jia, Kaipeng and Peng, Jiaxin and Huang, Yuyao and Xiong, Sishi and Jiang, Zhuo and Yu, Kaidong and Hu, Xiaohui and Yao, Fubei and Fang, Ruiyu and Jiang, Zhuoru and Song, Ruiting and Xie, Qiyi and Xue, Rui and He, Xuewei and Xue, Yanlei and Yuan, Zhu and Zhang, Zhaoxi and Huang, Zilu and Wang, Shiquan and Wang, Xin and Wu, Hanming and Wang, Mingyuan and Zhan, Xufeng and Sun, Yuhan and Xing, Zhaohu and Jiang, Yuhao and Yang, Bingkai and Song, Shuangyong and Li, Yongxiang and He, Zhongjiang and Li, Xuelong},
  year = {2025},
  eprint = {2507.18013},
  archivePrefix = {arXiv},
  primaryClass = {cs.CL},
  url = {https://arxiv.org/abs/2507.18013},
  note = {arXiv preprint arXiv:2507.18013}
}

@misc{deepseek-v2,
  title = {DeepSeek-V2: A Strong, Economical, and Efficient Mixture-of-Experts Language Model},
  author = {{DeepSeek-AI} and Aixin Liu and Bei Feng and Bin Wang and Bingxuan Wang and Bo Liu and
            Chenggang Zhao and Chengqi Dengr and Chong Ruan and Damai Dai and Daya Guo and Dejian Yang
            and Deli Chen and Dongjie Ji and Erhang Li and Fangyun Lin and Fuli Luo and Guangbo Hao and
            Guanting Chen and Guowei Li and H. Zhang and Hanwei Xu and Hao Yang and Haowei Zhang and
            Honghui Ding and Huajian Xin and Huazuo Gao and Hui Li and Hui Qu and J. L. Cai and Jian
            Liang and Jianzhong Guo and Jiaqi Ni and Jiashi Li and Jin Chen and Jingyang Yuan and Junjie
            Qiu and Junxiao Song and Kai Dong and Kaige Gao and Kang Guan and Lean Wang and Lecong Zhang
            and Lei Xu and Leyi Xia and Liang Zhao and Liyue Zhang and Meng Li and Miaojun Wang and
            Mingchuan Zhang and Minghua Zhang and Minghui Tang and Mingming Li and Ning Tian and Panpan
            Huang and Peiyi Wang and Peng Zhang and Qihao Zhu and Qinyu Chen and Qiushi Du and R. J.
            Chen and R. L. Jin and Ruiqi Ge and Ruizhe Pan and Runxin Xu and Ruyi Chen and S. S. Li and
            Shanghao Lu and Shangyan Zhou and Shanhuang Chen and Shaoqing Wu and Shengfeng Ye and
            Shirong Ma and Shiyu Wang and Shuang Zhou and Shuiping Yu and Shunfeng Zhou and Size Zheng
            and T. Wang and Tian Pei and Tian Yuan and Tianyu Sun and W. L. Xiao and Wangding Zeng and
            Wei An and Wen Liu and Wenfeng Liang and Wenjun Gao and Wentao Zhang and X. Q. Li and
            Xiangyue Jin and Xianzu Wang and Xiao Bi and Xiaodong Liu and Xiaohan Wang and Xiaojin Shen
            and Xiaokang Chen and Xiaosha Chen and Xiaotao Nie and Xiaowen Sun and Xiaoxiang Wang and
            Xin Liu and Xin Xie and Xingkai Yu and Xinnan Song and Xinyi Zhou and Xinyu Yang and Xuan Lu
            and Xuecheng Su and Y. Wu and Y. K. Li and Y. X. Wei and Y. X. Zhu and Yanhong Xu and
            Yanping Huang and Yao Li and Yao Zhao and Yaofeng Sun and Yaohui Li and Yaohui Wang and Yi
            Zheng and Yichao Zhang and Yiliang Xiong and Yilong Zhao and Ying He and Ying Tang and Yishi
            Piao and Yixin Dong and Yixuan Tan and Yiyuan Liu and Yongji Wang and Yongqiang Guo and
            Yuchen Zhu and Yuduan Wang and Yuheng Zou and Yukun Zha and Yunxian Ma and Yuting Yan and
            Yuxiang You and Yuxuan Liu and Z. Z. Ren and Zehui Ren and Zhangli Sha and Zhe Fu and Zhen
            Huang and Zhen Zhang and Zhenda Xie and Zhewen Hao and Zhihong Shao and Zhiniu Wen and
            Zhipeng Xu and Zhongyu Zhang and Zhuoshu Li and Zihan Wang and Zihui Gu and Zilin Li and
            Ziwei Xie},
  year = {2024},
  eprint = {2405.04434},
  archivePrefix = {arXiv},
  primaryClass = {cs.CL},
  url = {https://arxiv.org/abs/2405.04434},
  note = {arXiv preprint arXiv:2405.04434}
}

@misc{deepseek-v3,
  title = {DeepSeek-V3 Technical Report},
  author = {{DeepSeek-AI} and Aixin Liu and Bei Feng and Bing Xue and Bingxuan Wang and Bochao Wu and
            Chengda Lu and Chenggang Zhao and Chengqi Deng and Chenyu Zhang and Chong Ruan and Damai Dai
            and Daya Guo and Dejian Yang and Deli Chen and Dongjie Ji and Erhang Li and Fangyun Lin and
            Fucong Dai and Fuli Luo and Guangbo Hao and Guanting Chen and Guowei Li and H. Zhang and Han
            Bao and Hanwei Xu and Haocheng Wang and Haowei Zhang and Honghui Ding and Huajian Xin and
            Huazuo Gao and Hui Li and Hui Qu and J. L. Cai and Jian Liang and Jianzhong Guo and Jiaqi Ni
            and Jiashi Li and Jiawei Wang and Jin Chen and Jingchang Chen and Jingyang Yuan and Junjie
            Qiu and Junlong Li and Junxiao Song and Kai Dong and Kai Hu and Kaige Gao and Kang Guan and
            Kexin Huang and Kuai Yu and Lean Wang and Lecong Zhang and Lei Xu and Leyi Xia and Liang
            Zhao and Litong Wang and Liyue Zhang and Meng Li and Miaojun Wang and Mingchuan Zhang and
            Minghua Zhang and Minghui Tang and Mingming Li and Ning Tian and Panpan Huang and Peiyi Wang
            and Peng Zhang and Qiancheng Wang and Qihao Zhu and Qinyu Chen and Qiushi Du and R. J. Chen
            and R. L. Jin and Ruiqi Ge and Ruisong Zhang and Ruizhe Pan and Runji Wang and Runxin Xu and
            Ruoyu Zhang and Ruyi Chen and S. S. Li and Shanghao Lu and Shangyan Zhou and Shanhuang Chen
            and Shaoqing Wu and Shengfeng Ye and Shengfeng Ye and Shirong Ma and Shiyu Wang and Shuang
            Zhou and Shuiping Yu and Shunfeng Zhou and Shuting Pan and T. Wang and Tao Yun and Tian Pei
            and Tianyu Sun and W. L. Xiao and Wangding Zeng and Wanjia Zhao and Wei An and Wen Liu and
            Wenfeng Liang and Wenjun Gao and Wenqin Yu and Wentao Zhang and X. Q. Li and Xiangyue Jin
            and Xianzu Wang and Xiao Bi and Xiaodong Liu and Xiaohan Wang and Xiaojin Shen and Xiaokang
            Chen and Xiaokang Zhang and Xiaosha Chen and Xiaotao Nie and Xiaowen Sun and Xiaoxiang Wang
            and Xin Cheng and Xin Liu and Xin Xie and Xingchao Liu and Xingkai Yu and Xinnan Song and
            Xinxia Shan and Xinyi Zhou and Xinyu Yang and Xinyuan Li and Xuecheng Su and Xuheng Lin and
            Y. K. Li and Y. Q. Wang and Y. X. Wei and Y. X. Zhu and Yang Zhang and Yanhong Xu and
            Yanhong Xu and Yanping Huang and Yao Li and Yao Zhao and Yaofeng Sun and Yaohui Li and
            Yaohui Wang and Yi Yu and Yi Zheng and Yichao Zhang and Yifan Shi and Yiliang Xiong and Ying
            He and Ying Tang and Yishi Piao and Yisong Wang and Yixuan Tan and Yiyang Ma and Yiyuan Liu
            and Yongqiang Guo and Yu Wu and Yuan Ou and Yuchen Zhu and Yuduan Wang and Yue Gong and
            Yuheng Zou and Yujia He and Yukun Zha and Yunfan Xiong and Yunxian Ma and Yuting Yan and
            Yuxiang Luo and Yuxiang You and Yuxuan Liu and Yuyang Zhou and Z. F. Wu and Z. Z. Ren and
            Zehui Ren and Zhangli Sha and Zhe Fu and Zhean Xu and Zhen Huang and Zhen Zhang and Zhenda
            Xie and Zhengyan Zhang and Zhewen Hao and Zhibin Gou and Zhicheng Ma and Zhigang Yan and
            Zhihong Shao and Zhipeng Xu and Zhiyu Wu and Zhongyu Zhang and Zhuoshu Li and Zihui Gu and
            Zijia Zhu and Zijun Liu and Zilin Li and Ziwei Xie and Ziyang Song and Ziyi Gao and Zizheng
            Pan},
  year = {2024},
  eprint = {2412.19437},
  archivePrefix = {arXiv},
  primaryClass = {cs.CL},
  url = {https://arxiv.org/abs/2412.19437},
  note = {arXiv preprint arXiv:2412.19437}
}

@misc{qwen3,
  title = {Qwen3 Technical Report},
  author = {An Yang and Anfeng Li and Baosong Yang and Beichen Zhang and Binyuan Hui and Bo Zheng and
            Bowen Yu and Chang Gao and Chengen Huang and Chenxu Lv and Chujie Zheng and Dayiheng Liu and
            Fan Zhou and Fei Huang and Feng Hu and Hao Ge and Haoran Wei and Huan Lin and Jialong Tang
            and Jian Yang and Jianhong Tu and Jianwei Zhang and Jianxin Yang and Jiaxi Yang and Jing
            Zhou and Jingren Zhou and Junyang Lin and Kai Dang and Keqin Bao and Kexin Yang and Le Yu
            and Lianghao Deng and Mei Li and Mingfeng Xue and Mingze Li and Pei Zhang and Peng Wang and
            Qin Zhu and Rui Men and Ruize Gao and Shixuan Liu and Shuang Luo and Tianhao Li and Tianyi
            Tang and Wenbiao Yin and Xingzhang Ren and Xinyu Wang and Xinyu Zhang and Xuancheng Ren and
            Yang Fan and Yang Su and Yichang Zhang and Yinger Zhang and Yu Wan and Yuqiong Liu and Zekun
            Wang and Zeyu Cui and Zhenru Zhang and Zhipeng Zhou and Zihan Qiu},
  year = {2025},
  eprint = {2505.09388},
  archivePrefix = {arXiv},
  primaryClass = {cs.CL},
  url = {https://arxiv.org/abs/2505.09388},
  note = {arXiv preprint arXiv:2505.09388}
}

@misc{mindspore-framework,
  title = {{MindSpore}},
  author = {{MindSpore Team}},
  year = {n.d.-b},
  howpublished = {Official software repository},
  url = {https://github.com/mindspore-ai/mindspore},
  note = {Accessed November 12, 2026}
}

@misc{deepseekai2025deepseekv32pushingfrontieropen,
      title={DeepSeek-V3.2: Pushing the Frontier of Open Large Language Models}, 
      author={DeepSeek-AI and Aixin Liu and Aoxue Mei and Bangcai Lin and Bing Xue and Bingxuan Wang and Bingzheng Xu and Bochao Wu and Bowei Zhang and Chaofan Lin and Chen Dong and Chengda Lu and Chenggang Zhao and Chengqi Deng and Chenhao Xu and Chong Ruan and Damai Dai and Daya Guo and Dejian Yang and Deli Chen and Erhang Li and Fangqi Zhou and Fangyun Lin and Fucong Dai and Guangbo Hao and Guanting Chen and Guowei Li and H. Zhang and Hanwei Xu and Hao Li and Haofen Liang and Haoran Wei and Haowei Zhang and Haowen Luo and Haozhe Ji and Honghui Ding and Hongxuan Tang and Huanqi Cao and Huazuo Gao and Hui Qu and Hui Zeng and Jialiang Huang and Jiashi Li and Jiaxin Xu and Jiewen Hu and Jingchang Chen and Jingting Xiang and Jingyang Yuan and Jingyuan Cheng and Jinhua Zhu and Jun Ran and Junguang Jiang and Junjie Qiu and Junlong Li and Junxiao Song and Kai Dong and Kaige Gao and Kang Guan and Kexin Huang and Kexing Zhou and Kezhao Huang and Kuai Yu and Lean Wang and Lecong Zhang and Lei Wang and Liang Zhao and Liangsheng Yin and Lihua Guo and Lingxiao Luo and Linwang Ma and Litong Wang and Liyue Zhang and M. S. Di and M. Y Xu and Mingchuan Zhang and Minghua Zhang and Minghui Tang and Mingxu Zhou and Panpan Huang and Peixin Cong and Peiyi Wang and Qiancheng Wang and Qihao Zhu and Qingyang Li and Qinyu Chen and Qiushi Du and Ruiling Xu and Ruiqi Ge and Ruisong Zhang and Ruizhe Pan and Runji Wang and Runqiu Yin and Runxin Xu and Ruomeng Shen and Ruoyu Zhang and S. H. Liu and Shanghao Lu and Shangyan Zhou and Shanhuang Chen and Shaofei Cai and Shaoyuan Chen and Shengding Hu and Shengyu Liu and Shiqiang Hu and Shirong Ma and Shiyu Wang and Shuiping Yu and Shunfeng Zhou and Shuting Pan and Songyang Zhou and Tao Ni and Tao Yun and Tian Pei and Tian Ye and Tianyuan Yue and Wangding Zeng and Wen Liu and Wenfeng Liang and Wenjie Pang and Wenjing Luo and Wenjun Gao and Wentao Zhang and Xi Gao and Xiangwen Wang and Xiao Bi and Xiaodong Liu and Xiaohan Wang and Xiaokang Chen and Xiaokang Zhang and Xiaotao Nie and Xin Cheng and Xin Liu and Xin Xie and Xingchao Liu and Xingkai Yu and Xingyou Li and Xinyu Yang and Xinyuan Li and Xu Chen and Xuecheng Su and Xuehai Pan and Xuheng Lin and Xuwei Fu and Y. Q. Wang and Yang Zhang and Yanhong Xu and Yanru Ma and Yao Li and Yao Li and Yao Zhao and Yaofeng Sun and Yaohui Wang and Yi Qian and Yi Yu and Yichao Zhang and Yifan Ding and Yifan Shi and Yiliang Xiong and Ying He and Ying Zhou and Yinmin Zhong and Yishi Piao and Yisong Wang and Yixiao Chen and Yixuan Tan and Yixuan Wei and Yiyang Ma and Yiyuan Liu and Yonglun Yang and Yongqiang Guo and Yongtong Wu and Yu Wu and Yuan Cheng and Yuan Ou and Yuanfan Xu and Yuduan Wang and Yue Gong and Yuhan Wu and Yuheng Zou and Yukun Li and Yunfan Xiong and Yuxiang Luo and Yuxiang You and Yuxuan Liu and Yuyang Zhou and Z. F. Wu and Z. Z. Ren and Zehua Zhao and Zehui Ren and Zhangli Sha and Zhe Fu and Zhean Xu and Zhenda Xie and Zhengyan Zhang and Zhewen Hao and Zhibin Gou and Zhicheng Ma and Zhigang Yan and Zhihong Shao and Zhixian Huang and Zhiyu Wu and Zhuoshu Li and Zhuping Zhang and Zian Xu and Zihao Wang and Zihui Gu and Zijia Zhu and Zilin Li and Zipeng Zhang and Ziwei Xie and Ziyi Gao and Zizheng Pan and Zongqing Yao and Bei Feng and Hui Li and J. L. Cai and Jiaqi Ni and Lei Xu and Meng Li and Ning Tian and R. J. Chen and R. L. Jin and S. S. Li and Shuang Zhou and Tianyu Sun and X. Q. Li and Xiangyue Jin and Xiaojin Shen and Xiaosha Chen and Xinnan Song and Xinyi Zhou and Y. X. Zhu and Yanping Huang and Yaohui Li and Yi Zheng and Yuchen Zhu and Yunxian Ma and Zhen Huang and Zhipeng Xu and Zhongyu Zhang and Dongjie Ji and Jian Liang and Jianzhong Guo and Jin Chen and Leyi Xia and Miaojun Wang and Mingming Li and Peng Zhang and Ruyi Chen and Shangmian Sun and Shaoqing Wu and Shengfeng Ye and T. Wang and W. L. Xiao and Wei An and Xianzu Wang and Xiaowen Sun and Xiaoxiang Wang and Ying Tang and Yukun Zha and Zekai Zhang and Zhe Ju and Zhen Zhang and Zihua Qu},
      year={2025},
      eprint={2512.02556},
      archivePrefix={arXiv},
      primaryClass={cs.CL},
      url={https://arxiv.org/abs/2512.02556}, 
}

@misc{bai2026indexcacheacceleratingsparseattention,
      title={IndexCache: Accelerating Sparse Attention via Cross-Layer Index Reuse}, 
      author={Yushi Bai and Qian Dong and Ting Jiang and Xin Lv and Zhengxiao Du and Aohan Zeng and Jie Tang and Juanzi Li},
      year={2026},
      eprint={2603.12201},
      archivePrefix={arXiv},
      primaryClass={cs.CL},
      url={https://arxiv.org/abs/2603.12201}, 
}

@misc{telechat3moe2025,
  title = {Training Report of {TeleChat3-MoE}},
  author = {Xinzhang Liu and Chao Wang and Zhihao Yang and Zhuo Jiang and Xuncheng Zhao and Haoran Wang
            and Lei Li and Dongdong He and Luobin Liu and Kaizhe Yuan and Han Gao and Zihan Wang and
            Yitong Yao and Sishi Xiong and Wenmin Deng and Haowei He and Kaidong Yu and Yu Zhao and
            Ruiyu Fang and Yuhao Jiang and Yingyan Li and Xiaohui Hu and Xi Yu and Jingqi Li and Yanwei
            Liu and Qingli Li and Xinyu Shi and Junhao Niu and Chengnuo Huang and Yao Xiao and Ruiwen
            Wang and Fengkai Li and Luwen Pu and Kaipeng Jia and Fubei Yao and Yuyao Huang and Xuewei He
            and Zhuoru Jiang and Ruiting Song and Rui Xue and Qiyi Xie and Jie Zhang and Zilu Huang and
            Zhaoxi Zhang and Zhilong Lu and Yanhan Zhang and Yin Zhang and Yanlei Xue and Zhu Yuan and
            Teng Su and Xin Jiang and Shuangyong Song and Yongxiang Li and Xuelong Li},
  year = {2025},
  eprint = {2512.24157},
  archivePrefix = {arXiv},
  url = {https://arxiv.org/abs/2512.24157},
  note = {arXiv preprint arXiv:2512.24157}
}

@misc{hunyuan2026a13b,
  title = {{Hunyuan-A13B} Technical Report},
  author = {{Tencent Hunyuan Team} and Ao Liu and Botong Zhou and Can Xu and Chayse Zhou and ChenChen
            Zhang and Chengcheng Xu and Chenhao Wang and Decheng Wu and Dengpeng Wu and Dian Jiao and
            Dong Du and Dong Wang and Feng Zhang and Fengzong Lian and Guanghui Xu and Guanwei Zhang and
            Hai Wang and Haipeng Luo and Han Hu and Huilin Xu and Jiajia Wu and Jianchen Zhu and
            Jianfeng Yan and Jiaqi Zhu and Jihong Zhang and Jinbao Xue and Jun Xia and Junqiang Zheng
            and Kai Liu and Kai Zhang and Kai Zheng and Kejiao Li and Keyao Wang and Lan Jiang and Lixin
            Liu and Lulu Wu and Mengyuan Huang and Peijie Yu and Peiqi Wang and Qian Wang and Qianbiao
            Xiang and Qibin Liu and Qingfeng Sun and Richard Guo and Ruobing Xie and Saiyong Yang and
            Shaohua Chen and Shihui Hu and Shuai Li and Shuaipeng Li and Shuang Chen and Suncong Zheng
            and Tao Yang and Tian Zhang and Tinghao Yu and Weidong Han and Weijie Liu and Weijin Zhou
            and Weikang Wang and Wesleye Chen and Xiao Feng and Xiaoqin Ren and Xingwu Sun and Xiong
            Kuang and Xuemeng Huang and Xun Cao and Yanfeng Chen and Yang Du and Zhen Yang and Yangyu
            Tao and Yaping Deng and Yi Shen and Yigeng Hong and Yiqi Chen},
  year = {2026},
  eprint = {2609.27284},
  archivePrefix = {arXiv},
  url = {https://arxiv.org/abs/2609.27284},
  note = {arXiv preprint arXiv:2609.27284}
}

@misc{kimi2026k3,
  title = {{Kimi K3}: Open Frontier Intelligence},
  author = {{Kimi Team} and Tongtong Bai and Yifan Bai and Yiping Bao and M. C. and Jianfeng Cai and
            Xinyuan Cai and Peizhou Cao and Yuxuan Cao and Ziwei Chai and Y. Charles and H. S. Che and
            Guanduo Chen and Guangyu Chen and Guanzheng Chen and Huarong Chen and Jia Chen and Jianlong
            Chen and Jun Chen and Kexin Chen and Peng Chen and Ruijue Chen and Wentao Chen and Xin Chen
            and Yang Chen and Yanru Chen and Yifei Chen and Yingjiang Chen and Yuankun Chen and Yujie
            Chen and Yutian Chen and Zhirong Chen and Dazhi Cheng and Yean Cheng and Jialei Cui and
            Jingbing Cui and Anqi Dai and Jiaqi Deng and Hao Ding and Rui Ding and Shaofeng Ding and
            Mengfan Dong and Mengnan Dong and Yuhao Dong and Yuxin Dong and Angang Du and Chenzhuang Du
            and Dikang Du and Jusen Du and Yulun Du and Yu Fan and Jing Feng and Qiulin Feng and Yichen
            Feng and Kelin Fu and Qiang Fu and Fuxuan Gao and Hongcheng Gao and Jingyue Gao and Tong Gao
            and Weijia Gao and Shangyi Geng and Jie Gong and Linhu Gong and Shengao Gong and Xiaochen
            Gong and Qizheng Gu and Yicheng Gu and Shuhao Guan and Haiqing Guo and Shiqi Guo and Xiang
            Guo and Zhengyan Guo and Beixi Hao and Wenxin Hao and Xiaoru Hao and Dailan He and Haotian
            He and Lehan He and Qi He and Weiran He and Xinran He and Xinyi He and Yibo He and Yunjia He
            and Chao Hong and Tiange Hong and Hao Hu and Jiaxi Hu and Ruikun Hu and Weiming Hu and
            Yangyang Hu and Zhenxing Hu and Liang Hua and Jinbin Huang and Ke Huang and Ruiyuan Huang
            and Siying Huang and Weixiao Huang and Yan Huang and Zhengjie Huang and Zhiqi Huang and
            Yulong Hui and Chaobo Jia and Yutong Jiang and Zhejun Jiang and Zuoyou Jiang and Wenyi Jin
            and Xinyi Jin and Yu Jing and Huanjun Kong and Guokun Lai and Aidi Li and Cheng Li and
            Chengyuan Li and Cong Li and Fang Li and Guanyu Li and Haoyang Li and Jia Li and Junxiong Li
            and Lei Li and Letian Li and Lincan Li and Weihong Li and Wentao Li and Xintong Li and Yang
            Li and Yishen Li and Yiwei Li and Yuxiao Li and Zhaowei Li and Zhaoxi Li and Zheming Li and
            Zhengxiao Li and Zhiyuan Li and Jiawei Lin and Xiaohan Lin and Yibo Lin and Zichao Lin and
            Ziyan Lin and Bill Liu and Boxiao Liu and Chuan Liu and Liang Liu and Shaowei Liu and
            Shudong Liu and Shuran Liu and Tianwei Liu and Weizhou Liu and Yangyang Liu and Yanming Liu
            and Yibo Liu and Yipeng Liu and Zhengying Liu and Zhiheng Liu and Enzhe Lu and Haoyu Lu and
            Linqiang Lu and Tingzhan Lu and Zhiyuan Lu and Aotian Luo and G. Luo and Junyu Luo and Yifan
            Luo and B. Lyu and Wenzhou Lyu and Shaoguang Mao and Yuan Mei and Xin Men and Minqing Ni and
            Yixuan Niu and Siyuan Pan and Shujun Peng and Zhangyang Qi and Ruoyu Qin and ZeChao Qin and
            Zeyu Qin and Haiquan Qiu and Jianxin Qiu and Jiezhong Qiu and Bowen Qu and Yuhao Qu and Zeyu
            Shang and Youbo Shao and Han Shen and Jincheng Shi and Juanfeng Shi and Lidong Shi and
            Shengyuan Shi and Wingchun Siu and Pengwei Song and Xiaoxi Song and Jianlin Su and Yunfeng
            Su and Zhaochen Su and Lin Sui and Jingsong Sun and Junyao Sun and Shaoning Sun and Shuzhe
            Sun and Tongyu Sun and Yujun Sun and Yunpeng Tai and Chuning Tang and Heyi Tang and Sirui
            Tang and Zecheng Tang and Chaoran Tian and Rongpeng Tian and Yu Tian and Wei Tu and Chensi
            Wang and Chuang Wang and Chunjie Wang and Dinglu Wang and Feng Wang and Hailong Wang and
            Haiming Wang and Hao Wang and Hao Wang and Huaqing Wang and Hui Wang and Jiayi Wang and
            Jinglong Wang and Jinhong Wang and Jiuzheng Wang and Linian Wang and Shaobo Wang and Shenzhi
            Wang and Shuyi Wang and Si Wang and Siyuan Wang and Tianfu Wang and Wenjue Wang and Xingran
            Wang and Xinmei Wang and Xinyuan Wang and Xusheng Wang and Yalin Wang and Yangkun Wang and
            Yao Wang and Yaoyu Wang and Yejie Wang and Yiqin Wang and Yucheng Wang and Yuzhi Wang and
            Zhaoji Wang and Zhaowei Wang and Zhengtao Wang and Zhenhao Wang and Zhongsheng Wang and
            Zifan Wang and Chu Wei and Ming Wei and Shouxin Wei and Zichen Wen and Fan Wu and Haoning Wu
            and Rucong Wu and Wenhao Wu and Xiaoxue Wu and Yingcong Wu and Yongqi Wu and Yuxin Wu and
            Zijian Wu and Xinglang Xian and Chenxuan Xiang and Yuye Xiang and Bocheng Xiao and Chenjun
            Xiao and Xin Xiao and Jin Xie and Xiaotong Xie and Yifeng Xie and Zhe Xie and Bowei Xing and
            Yiming Xiong and Baosheng Xu and Boyu Xu and Jiale Xu and Jianfan Xu and Jing Xu and Jinjing
            Xu and L. H. Xu and Qingtao Xu and Shuyao Xu and Suting Xu and Tiantian Xu and Tianxiang Xu
            and Weixin Xu and Xinran Xu and Yangchuan Xu and Ye Xu and Yueni Xu and Ziyao Xu and Haonan
            Xue and Junjie Yan and Yaoyao Yan and Fan Yang and Guangyao Yang and Hao Yang and Junwei
            Yang and Ruoyu Yang and Wenjie Yang and Xiaofei Yang and Xinyu Yang and Yi Yang and Yiling
            Yang and Ying Yang and Yuchen Yang and Zhen Yang and Zhilin Yang and Zian Yang and Zuhao
            Yang and Haotian Yao and Dan Ye and Haoran Ye and Wenjie Ye and Zhanbo Ye and Bohong Yin and
            Haoxiang Yin and Xietong Yin and Chengzhen Yu and Haozhen Yu and Longhui Yu and Shengnan Yu
            and Shuying Yu and Tianxiang Yu and Enming Yuan and Mengjie Yuan and Tongtian Yue and Wei
            Yue and Yang Yue and Dunyuan Zha and Haobing Zhan and B. H. Zhang and Dehao Zhang and Fei
            Zhang and Hao Zhang and Haoyuan Zhang and Huanyu Zhang and Jiapei Zhang and Jiaxuan Zhang
            and Jin Zhang and Kaiyi Zhang and Miaozhen Zhang and Puqi Zhang and Qinglei Zhang and Rong
            Zhang and Rui Zhang and Shaoshuai Zhang and Shiyi Zhang and Xiaobin Zhang and Xiaoyun Zhang
            and Y. Zhang and Yangkun Zhang and Ye Zhang and Yichi Zhang and Yikun Zhang and Yizhi Zhang
            and Yongting Zhang and Yu Zhang and Yutao Zhang and Yutong Zhang and Zheng Zhang and Zijing
            Zhang and Bin Zhao and Chenguang Zhao and Feifan Zhao and Jinglun Zhao and Jinxiang Zhao and
            Shuai Zhao and Wenshuo Zhao and Xiangyu Zhao and Xuanle Zhao and Yikai Zhao and Zijia Zhao
            and Haozhi Zheng and Huabin Zheng and Ruihan Zheng and Shaojie Zheng and Tengyang Zheng and
            Haofeng Zhong and Lei Zhong and Longguang Zhong and M. Zhou and Qiankang Zhou and Runjie
            Zhou and Ruozhang Zhou and Xinyu Zhou and Yiqiao Zhou and Zaida Zhou and Jinguo Zhu and Liya
            Zhu and Xinhao Zhu and Yangjunfeng Zhu and Yuxuan Zhu and Zhen Zhu and Chen Zhuang and Weiyu
            Zhuang and Xinxing Zu},
  year = {2026},
  eprint = {2607.24653},
  archivePrefix = {arXiv},
  url = {https://arxiv.org/abs/2607.24653},
  note = {arXiv preprint arXiv:2607.24653}
}

@misc{glm2026five,
  title = {{GLM-5}: from Vibe Coding to Agentic Engineering},
  author = {{GLM-5-Team} and Aohan Zeng and Xin Lv and Zhenyu Hou and Zhengxiao Du and Qinkai Zheng and
            Bin Chen and Da Yin and Chendi Ge and Chenghua Huang and Chengxing Xie and Chenzheng Zhu and
            Congfeng Yin and Cunxiang Wang and Gengzheng Pan and Hao Zeng and Haoke Zhang and Haoran
            Wang and Huilong Chen and Jiajie Zhang and Jian Jiao and Jiaqi Guo and Jingsen Wang and
            Jingzhao Du and Jinzhu Wu and Kedong Wang and Lei Li and Lin Fan and Lucen Zhong and Mingdao
            Liu and Mingming Zhao and Pengfan Du and Qian Dong and Rui Lu and Shuang-Li and Shulin Cao
            and Song Liu and Ting Jiang and Xiaodong Chen and Xiaohan Zhang and Xuancheng Huang and
            Xuezhen Dong and Yabo Xu and Yao Wei and Yifan An and Yilin Niu and Yitong Zhu and Yuanhao
            Wen and Yukuo Cen and Yushi Bai and Zhongpei Qiao and Zihan Wang and Zikang Wang and Zilin
            Zhu and Ziqiang Liu and Zixuan Li and Bojie Wang and Bosi Wen and Can Huang and Changpeng
            Cai and Chao Yu and Chen Li and Chengwei Hu and Chenhui Zhang and Dan Zhang and Daoyan Lin
            and Dayong Yang and Di Wang and Ding Ai and Erle Zhu and Fangzhou Yi and Feiyu Chen and
            Guohong Wen and Hailong Sun and Haisha Zhao and Haiyi Hu and Hanchen Zhang and Hanrui Liu
            and Hanyu Zhang and Hao Peng and Hao Tai and Haobo Zhang and He Liu and Hongwei Wang and
            Hongxi Yan and Hongyu Ge and Huan Liu and Huanpeng Chu and Jia'ni Zhao and Jiachen Wang and
            Jiajing Zhao and Jiamin Ren and Jiapeng Wang and Jiaxin Zhang and Jiayi Gui and Jiayue Zhao
            and Jijie Li and Jing An and Jing Li and Jingwei Yuan and Jinhua Du and Jinxin Liu and
            Junkai Zhi and Junwen Duan and Kaiyue Zhou and Kangjian Wei and Ke Wang and Keyun Luo and
            Laiqiang Zhang and Leigang Sha and Liang Xu and Lindong Wu and Lintao Ding and Lu Chen and
            Minghao Li and Nianyi Lin and Pan Ta and Qiang Zou and Rongjun Song and Ruiqi Yang and
            Shangqing Tu and Shangtong Yang and Shaoxiang Wu and Shengyan Zhang and Shijie Li and Shuang
            Li and Shuyi Fan and Wei Qin and Wei Tian and Weining Zhang and Wenbo Yu and Wenjie Liang
            and Xiang Kuang and Xiangmeng Cheng and Xiangyang Li and Xiaoquan Yan and Xiaowei Hu and
            Xiaoying Ling and Xing Fan and Xingye Xia and Xinyuan Zhang and Xinze Zhang and Xirui Pan
            and Xu Zou and Xunkai Zhang and Yadi Liu and Yandong Wu and Yanfu Li and Yidong Wang and
            Yifan Zhu and Yijun Tan and Yilin Zhou and Yiming Pan and Ying Zhang and Yinpei Su and
            Yipeng Geng and Yong Yan and Yonglin Tan and Yuean Bi and Yuhan Shen and Yuhao Yang and
            Yujiang Li and Yunan Liu and Yunqing Wang and Yuntao Li and Yurong Wu and Yutao Zhang and
            Yuxi Duan and Yuxuan Zhang and Zezhen Liu and Zhengtao Jiang and Zhenhe Yan and Zheyu Zhang
            and Zhixiang Wei and Zhuo Chen and Zhuoer Feng and Zijun Yao and Ziwei Chai and Ziyuan Wang
            and Zuzhou Zhang and Bin Xu and Minlie Huang and Hongning Wang and Juanzi Li and Yuxiao Dong
            and Jie Tang},
  year = {2026},
  eprint = {2602.15763},
  archivePrefix = {arXiv},
  url = {https://arxiv.org/abs/2602.15763},
  note = {arXiv preprint arXiv:2602.15763}
}

@misc{deepseek2026v4,
  title = {{DeepSeek-V4}: Towards Highly Efficient Million-Token Context Intelligence},
  author = {{DeepSeek-AI} and Anyi Xu and Bangcai Lin and Bing Xue and Bingxuan Wang and Bingzheng Xu
            and Bochao Wu and Bowei Zhang and Chaofan Lin and Chen Dong and Chenchen Ling and Chengda Lu
            and Chenggang Zhao and Chengqi Deng and Chengyu Hou and Chenhao Xu and Chenze Shao and Chong
            Ruan and Conner Sun and Damai Dai and Daya Guo and Dejian Yang and Deli Chen and Donghao Li
            and Dongjie Ji and Erhang Li and Fang Wei and Fangyun Lin and Fangzhou Yuan and Feiyu Xia
            and Fucong Dai and Guangbo Hao and Guanting Chen and Guoai Cao and Guolai Meng and Guowei Li
            and Han Yu and Han Zhang and Hanwei Xu and Hao Li and Haofen Liang and Haoling Zhang and
            Haoming Luo and Haoran Wei and Haotian Yuan and Haowei Zhang and Haowen Luo and Haoyu Chen
            and Haozhe Ji and Hengqing Zhang and Honghui Ding and Hongxuan Tang and Huanqi Cao and
            Huazuo Gao and Hui Qu and Hui Zeng and J Yang and JQ Zhu and Jia Luo and Jia Song and Jia Yu
            and Jialiang Huang and Jialu Cai and Jian Liang and Jiangting Zhou and Jiasheng Ye and
            Jiashi Li and Jiaxin Xu and Jiewen Hu and Jieyu Yang and Jin Chen and Jin Yan and Jingchang
            Chen and Jingli Zhou and Jingting Xiang and Jingyang Yuan and Jingyuan Cheng and Jingzi Zhou
            and Jinhua Zhu and Jiping Yu and Joseph Sun and Jun Ran and Junguang Jiang and Junjie Qiu
            and Junlong Li and Junmin Zheng and Junxiao Song and Kai Dong and Kaige Gao and Kang Guan
            and Kexing Zhou and Kezhao Huang and Kuai Yu and Lean Wang and Lecong Zhang and Lei Wang and
            Leyi Xia and Li Zhang and Liang Zhao and Lihua Guo and Lingxiao Luo and Linwang Ma and
            Linyan Zhu and Litong Wang and Liyu Cai and Liyue Zhang and Longhao Chen and MS Di and MY Xu
            and Max Mei and Miaojun Wang and Mingchuan Zhang and Minghua Zhang and Minghui Tang and
            Mingming Li and Mingxu Zhou and Minmin Han and Ning Wang and Panpan Huang and Panpan Wang
            and Peixin Cong and Peiyi Wang and Peng Zhang and Qiancheng Wang and Qihao Zhu and Qingyang
            Li and Qinyu Chen and Qiushi Du and Qiwei Jiang and Rui Tian and Ruifan Xu and Ruijie Lu and
            Ruiling Xu and Ruiqi Ge and Ruisong Zhang and Ruizhe Pan and Runji Wang and Runqian Chen and
            Runqiu Yin and Runxin Xu and Ruomeng Shen and Ruoyu Zhang and Ruyi Chen and SH Liu and
            Shanghao Lu and Shangmian Sun and Shangyan Zhou and Shanhuang Chen and Shaofei Cai and
            Shaoheng Nie and Shaoqing Wu and Shaoyuan Chen and Shengding Hu and Shengyu Liu and Shiqiang
            Hu and Shirong Ma and Shiyu Wang and Shuiping Yu and Shunfeng Zhou and Shuting Pan and
            Shuying Yu and Songyang Zhou and Tao Ni and Tao Yun and Tian Jin and Tian Pei and Tian Ye
            and Tianle Lin and Tianran Ji and Tianyi Cui and Tianyuan Yue and Tingting Yu and Tun Wang
            and W Zhang and WL Xiao and Wangding Zeng and Wei An and Weilin Zhao and Wen Liu and Wenfeng
            Liang and Wenjie Pang and Wenjing Luo and Wenjing Yao and Wenjun Gao and Wenkai Yang and
            Wenlve Huang and Wenqing Hou and Wentao Zhang and Wenting Ma and Xi Gao and Xiang He and
            Xiangwen Wang and Xianzu Wang and Xiao Bi and Xiaodong Liu and Xiaohan Wang and Xiaokang
            Chen and Xiaokang Zhang and Xiaotao Nie and Xiaowen Sun and Xiaoxiang Wang and Xin Cheng and
            Xin Liu and Xin Xie and Xingchao Liu and Xingchen Liu and Xingkai Yu and Xingyou Li and
            Xinyu Yang and Xinyu Zhang and Xu Chen and Xuanyu Wang and Xuecheng Su and Xueyin Chen and
            Xuheng Lin and Xuwei Fu and YC Yan and YQ Wang and YW Ma and Yanfeng Luo and Yang Zhang and
            Yanhong Xu and Yanru Ma and Yanwen Huang and Yao Li and Yao Li and Yao Xu and Yao Zhao and
            Yaofeng Sun and Yaohui Wang and Yi Qian and Yi Shao and Yi Yu and Yichao Zhang and Yifan
            Ding and Yifan Shi and Yijia Wu and Yiliang Xiong and Yiling Ma and Ying He and Ying Tang
            and Ying Zhou and Yingjia Luo and Yinmin Zhong and Yishi Piao and Yisong Wang and Yixiang
            Zhang and Yixiao Chen and Yixuan Tan and Yixuan Wei and Yiyang Ma and Yiyuan Liu and Yonglun
            Yang and Yongqiang Guo and Yongtong Wu and Yu Wu and YuKun Li and Yuan Cheng and Yuan Ou and
            Yuanfan Xu and Yuanhao Li and Yuduan Wang and Yuehan Yang and Yuer Xu and Yuhan Wu and Yuhao
            Meng and Yuheng Zou and Yukun Zha and Yunfan Xiong and Yupeng Chen and Yuping Lin and Yuqian
            Cao and Yuqian Wang and Yushun Zhang and Yuting Yan and Yutong Lin and Yuxian Gu and Yuxiang
            Luo and Yuxiang You and Yuxuan Liu and Yuxuan Zhou and Yuyang Zhou and Yuzhen Huang and ZF
            Wu and Zehao Wang and Zehua Zhao and Zehui Ren and Zekai Zhang and Zhangli Sha and Zhe Fu
            and Zhe Ju and Zhean Xu and Zhenda Xie and Zhengyan Zhang and Zheren Gao and Zhewen Hao and
            Zhibin Gou and Zhicheng Ma and Zhigang Yan and Zhihong Shao and Zhixian Huang and Zhixuan
            Chen and Zhiyu Wu and Zhizhou Ren and Zhongyu Wu and Zhuoshu Li and Zhuping Zhang and Zian
            Xu and Zihao Wang and Zihua Qu and Zihui Gu and Zijia Zhu and Zilin Li and Zipeng Zhang and
            Ziwei Xie and Ziyi Gao and Ziyi Wan and Zizheng Pan and Zongqing Yao},
  year = {2026},
  eprint = {2606.19348},
  archivePrefix = {arXiv},
  url = {https://arxiv.org/abs/2606.19348},
  note = {arXiv preprint arXiv:2606.19348}
}

@misc{zai2026glm52,
  title = {{GLM-5.2}: Built for Long-Horizon Tasks},
  author = {{Z.ai}},
  year = {2026},
  howpublished = {Official technical blog},
  url = {https://z.ai/blog/glm-5.2}
}

@misc{hyperparallel,
  title = {{HyperParallel}: A {SuperPod}-Affinity Training Acceleration Framework},
  author = {{MindSpore Team}},
  year = {n.d.-a},
  howpublished = {Official software repository},
  url = {https://github.com/mindspore-ai/hyper-parallel},
  note = {Accessed November 8, 2026}
}

@misc{zhao2023pytorch,
  title = {{PyTorch FSDP}: Experiences on Scaling Fully Sharded Data Parallel},
  author = {Yanli Zhao and Andrew Gu and Rohan Varma and Liang Luo and Chien-Chin Huang and Min Xu and
            Less Wright and Hamid Shojanazeri and Myle Ott and Sam Shleifer and Alban Desmaison and Can
            Balioglu and Pritam Damania and Bernard Nguyen and Geeta Chauhan and Yuchen Hao and Ajit
            Mathews and Shen Li},
  year = {2023},
  eprint = {2304.11277},
  archivePrefix = {arXiv},
  url = {https://arxiv.org/abs/2304.11277},
  note = {arXiv preprint arXiv:2304.11277}
}

@misc{liu2025muon,
  title = {{Muon} is Scalable for {LLM} Training},
  author = {Jingyuan Liu and Jianlin Su and Xingcheng Yao and Zhejun Jiang and Guokun Lai and Yulun Du
            and Yidao Qin and Weixin Xu and Enzhe Lu and Junjie Yan and Yanru Chen and Huabin Zheng and
            Yibo Liu and Shaowei Liu and Bohong Yin and Weiran He and Han Zhu and Yuzhi Wang and
            Jianzhou Wang and Mengnan Dong and Zheng Zhang and Yongsheng Kang and Hao Zhang and Xinran
            Xu and Yutao Zhang and Yuxin Wu and Xinyu Zhou and Zhilin Yang},
  year = {2025},
  eprint = {2502.16982},
  archivePrefix = {arXiv},
  url = {https://arxiv.org/abs/2502.16982},
  note = {arXiv preprint arXiv:2502.16982}
}

@article{hyperparallelfsdp2026,
  author  = {Mo Sun and Yifan Yao and Yanwei Liu and Luobin Liu and Zhenzhang Yang and Kaisheng Wang and
             Xiangyu Meng and Chen Li and Xizheng Pang and Huilan Li and Xinglei Xu and Yushi Cui and
             Xinyao Lin and Kaiqi Chen and Jie Zhang and Zeke Wang and Teng Su},
  title   = {{HyperParallel-FSDP}: Topology-Aware Fully Sharded Training with
             Layout-Driven Muon on {Ascend SuperPods}},
  journal = {arXiv preprint arXiv:2609.21594},
  year    = {2026},
  url     = {https://arxiv.org/abs/2609.21594}
}

@misc{linux-cgroup-v2,
  author = {Heo, Tejun},
  title = {Control Group v2},
  howpublished = {The Linux Kernel Documentation},
  year = {2015},
  url = {https://docs.kernel.org/admin-guide/cgroup-v2.html},
  note = {Continuously updated documentation. Accessed November 12, 2026}
}

@misc{linux-cpu-isolation,
  author = {{Kernel Development Community}},
  title = {{CPU} Isolation},
  howpublished = {The Linux Kernel Documentation},
  year = {n.d.},
  url = {https://docs.kernel.org/admin-guide/cpu-isolation.html},
  note = {Accessed November 12, 2026}
}

@misc{linux-sched-setaffinity,
  author = {{Linux man-pages Project}},
  title = {{sched\_setaffinity(2)}: Set and Get a Thread's {CPU} Affinity Mask},
  howpublished = {Linux man-pages 6.19},
  year = {2026},
  url = {https://man7.org/linux/man-pages/man2/sched_setaffinity.2.html},
  note = {Accessed November 12, 2026}
}

@misc{mindspore-msrun,
  author = {{MindSpore Team}},
  title = {{msrun} Launching},
  howpublished = {MindSpore Tutorials},
  year = {n.d.-c},
  url = {https://www.mindspore.cn/tutorials/en/master/parallel/msrun_launcher.html},
  note = {Accessed November 12, 2026}
}

@misc{xie2026mhc,
  title = {{mHC}: Manifold-Constrained Hyper-Connections},
  author = {Zhenda Xie and Yixuan Wei and Huanqi Cao and Chenggang Zhao and Chengqi Deng and Jiashi Li and Damai Dai and Huazuo Gao and Jiang Chang and Kuai Yu and Liang Zhao and Shangyan Zhou and Zhean Xu and Zhengyan Zhang and Wangding Zeng and Shengding Hu and Yuqing Wang and Jingyang Yuan and Lean Wang and Wenfeng Liang},
  year = {2026},
  eprint = {2512.24880},
  archivePrefix = {arXiv},
  primaryClass = {cs.CL},
  url = {https://arxiv.org/abs/2512.24880},
  note = {arXiv preprint arXiv:2512.24880}
}

@misc{xing4029b,
  author = {{XingChen-AGI}},
  title = {{Xing4.0-29B-A4B}},
  howpublished = {Official model card and weights, Hugging Face},
  year = {2026},
  url = {https://huggingface.co/XingChen-AGI/Xing4.0-29B-A4B},
  note = {Accessed November 14, 2026}
}
\bibliographystyle{iclr2024_conference}

\clearpage

\appendix

\section{Author List}
\label{app:author-list}

This work was initiated and led by Xingchen AGI Lab, China Telecom, in collaboration with Huawei and Harbin Institute of Technology (Shenzhen). The complete list of authors is as follows (authors are listed alphabetically by their first name):

Baotian Hu, Bingkai Yang, Boyang Wang, Changzai Pan, Chao Lyu, Chao Wang, Chen Li, Chenshuo Pan, Chenyu Lin, Chenyuan Zhang, Chongming Liu, Dawei Fan, Di Wu, Dong Guo, Dong Li, Dongfang Li, Enlin Liu, Fengkai Li, Fubei Yao, Gang Lin, Han Gao, Hanming Wu, Haowei He, Huapeng Liu, Hua Wang, Huinan Xu, Jiahong Xiong, Jiaxin Peng, Jianxiang Ma, Jianxing Zhuge, Jie Zhang, Ji Xiong, Jinglin Chen, Jingqi Li, Jingxuan Li, Jinhao Li, Jiwang Ye, Junchen Li, Junbiao Li, Junyang Wu, Kaidong Yu, Kaipeng Jia, Kehai Chen, Kuai Li, Laobin Liu, Lei Li, Libo Qin, Lingling Shi, Luwen Pu, Menghao Xu, Mengjiao Bao, Mengqi Chen, Min Zhang, Mingrui Wang, Naying Wang, Peicong Han, Peiming Guo, Pengfei Li, Pengyu Wang, Qiang Feng, Qi Wang, Qingjun Zhu, Qingli Li, Qitai Ding, Qiyi Xie, Quanchao Zhao, Rongbao Han, Rui Xue, Ruiyu Fang, Ruiting Song, Shijie Li, Shiquan Wang, Shuangyong Song, Sishi Xiong, Teng Su, Wankang Bao, Weijie Zhang, Wei Mao, Weiwei Fu, Weiyang Huang, Wenmin Deng, Wenqing Hu, Wujie Zhong, Xi Yu, Xiaohui Hu, Xiaomin Fang, Xiaomeng Huang, Xiaoqiong Shi, Xiaotong Shang, Xiaoyu Li, Xinglei Xu, Xin Jiang, Xin Wang, Xin Zhang, Xinzhang Liu, Xue Liu, Xuefeng Bai, Xuewei He, Yahan Li, Yanhan Zhang, Yanlei Xue, Yanzhang Wu, Yang Yu, Yefan Xiao, Yi Liu, Yin Zhang, Yingyan Li, Yitong Yao, Yixuan Li, Yongqiang Xia, Yongxiang Li, Yinzhou Zhu, Yu Zhao, Yuhao Jiang, Yunpeng Zhang, Yuting Lin, Yuxuan Wu, Yuyao Huang, Yuyi Wang, Zebo Peng, Zhenwu Pei, Zhenxiong Li, Zhao Wang, Zhaoxi Zhang, Zhihao Yang, Zhiqiang Dong, Zhiwei Yan, Zhilong Lu, Zhonghao Guo, Zhongheng Wu, Zhongjiang He, Zhongqiu Li, Zhu Yuan, Zhuoru Jiang, Zhuoyi Wei, Zijia Guo, Zilu Huang, Zhipeng Guo, Zirui Li, Zihan Wang.

\end{document}